\documentclass[runningheads]{llncs}

\newcommand{\bc}{\mathbf{c}}

\newcommand{\bI}{\mathbf{I}}

\newcommand{\bm}{\mathbf{m}}
\newcommand{\bn}{\mathbf{n}}\newcommand{\bN}{\mathbf{N}}

\newcommand{\bS}{\mathbf{S}}

\newcommand{\bx}{\mathbf{x}}

\newcommand{\cG}{\mathcal{G}}

\spnewtheorem*{nyquist}{Nyquist Conditions}{\bfseries}{\itshape}

\usepackage{eccv}

\usepackage{eccvabbrv}
\usepackage{bm}
\usepackage{multirow}
\usepackage{xurl}

\usepackage{graphicx}
\usepackage{booktabs}

\usepackage[accsupp]{axessibility}  %

\usepackage[pagebackref,breaklinks,colorlinks,citecolor=eccvblue]{hyperref}

\usepackage{orcidlink}
\hypersetup{
  pdftitle={Geometrically Accurate Feed-Forward Gaussian Splatting for Unbounded Scenes},
  pdfauthor={C\'esar D\'iaz Blanco, Zehao Yu, Haofei Xu, Andreas Geiger}
}

\begin{document}

\title{Geometrically Accurate Feed-Forward Gaussian Splatting for Unbounded Scenes}

\titlerunning{GAFFGSUS}

\author{C{\'e}sar D{\'i}az Blanco\inst{1} \and
Zehao Yu\inst{1,2} \and
Haofei Xu\inst{3,4} \and
Andreas Geiger\inst{1,2,3}}

\authorrunning{C.~D{\'i}az Blanco et al.}

\institute{%
$^{1}$University of T{\"u}bingen \quad
$^{2}$T{\"u}bingen AI Center \quad
$^{3}$KE:SAI \quad
$^{4}$ETH Zurich}

\maketitle

\begin{abstract}
While current feed-forward Gaussian splatting models achieve great visual quality given sparse-view images, their ability to accurately represent the scene's underlying geometry is underexplored, especially for unbounded outdoor scenes. To enable feed-forward Gaussian splatting models to achieve geometric accuracy, we introduce an edge-aware normal consistency loss that successfully supervises our encoder to predict Gaussians aligned to the scene's surfaces. For efficient training, we propose a fine-tuning strategy that leverages an encoder trained for photometric losses to work with new decoders specialized for geometric accuracy via a 3D smoothing filter. Moreover, our filter has ties to signal processing and projective camera models, ensuring greater visual quality when rendering at higher sampling rates. Our evaluation across several outdoor-scene datasets and ground-truth point clouds demonstrates that our method achieves state-of-the-art geometric accuracy by training solely on color images with direct photometric supervision and geometric self-supervision. Code is available \href{https://github.com/blancocd/GAFFGSUS}{here}.
\end{abstract}

\section{Introduction}
\label{sec:intro}
Understanding the 3D real world from 2D images is a fundamental challenge in computer vision. Foundational methods historically separated this task into two distinct stages: Structure-from-Motion (SfM) to recover camera poses and sparse features, followed by Multi-View Stereo (MVS) for dense reconstruction. More recently, differentiable representations such as Neural Radiance Fields~\cite{mildenhall2020nerf} and 3D Gaussian Splatting~\cite{kerbl3Dgaussians} have targeted dense reconstruction through the task of Novel View Synthesis (NVS). While these representations achieve unprecedented visual fidelity, they rely heavily on costly, per-scene iterative optimization, preventing their use in real-time applications.

Inspired by the zero-shot capabilities of large language models, recent work has shifted toward feed-forward architectures capable of dense scene reconstruction in a single forward pass. However, to date, these models have largely treated visual and geometric quality as mutually exclusive goals. Feed-forward Gaussian splatting models optimized primarily for photometric losses excel at novel view synthesis but fail to capture underlying scene geometry. Conversely, models targeting geometric accuracy are often restricted to highly constrained indoor environments or image pairs. The critical challenge of predicting explicit 3D representations with strict geometric consistency remains underexplored for unbounded, outdoor scenes.

In this paper, we bridge the gap between photometric and geometric objectives. Operating under the hypothesis that high visual fidelity translates to geometrically accurate representations when properly constrained, we present a feed-forward Gaussian splatting framework optimized for unbounded environments. We focus heavily on the extraction of high-quality meshes from sparse-view inputs (\eg, 8 or 16 images), given their widespread utility in downstream tasks. Leveraging the state-of-the-art ReSplat~\cite{xu2025resplat} architecture, we successfully transfer visual-quality priors to geometric accuracy. Our core contributions are threefold:
\begin{enumerate}
    \item 
    \textbf{Edge-Aware Normal Consistency Loss:} We introduce a self-supervised geometric loss tailored for sparse Gaussian representations. By scaling the loss with an image smoothness map, we successfully guide the encoder to predict Gaussians that lie flush against smooth scene surfaces.

    \item 
    \textbf{3D Smoothing Filter:} We adapt a 3D smoothing filter for the feed-forward setting to ensure predicted primitives respect the Nyquist-Shannon sampling theorem. This modification also eliminates degenerate subpixel projections and resolves numerical instabilities during training.

    \item
    \textbf{Robust Fine-Tuning Strategy:} As models grow deeper, repurposing trained encoders is essential for computational efficiency. We present a strategy to fine-tune an encoder, originally trained strictly on photometric losses with the gsplat~\cite{ye2025gsplat} decoder, using geometry-focused decoders like GGGS~\cite{Zhang2026GeometryGrounded} and Gaussian Opacity Fields~\cite{Yu2024GOF} in addition to our normal consistency loss.
\end{enumerate}

Extensive evaluations on datasets including Tanks and Temples~\cite{Knapitsch2017}, Mip-NeRF 360~\cite{barron2022mipnerf360}, and DL3DV~\cite{ling2024dl3dv} demonstrate that our approach yields representations that are both visually compelling and geometrically rigorous. Notably, our model achieves mesh precision competitive with costly per-scene optimization methods while operating in a single forward pass.

\section{Related Work}
\label{sec:related-work}

\textbf{Multi-View Depth Estimation.}
\label{subsubsec:mv-depth-estimation} 
Depth estimation is the task of predicting the distance from a camera to the environment it captures. Using multiple input images allows for capturing extended scenes and aids in disambiguating the predictions. Traditional methods~\cite{schoenberger2016mvs,schoenberger2016sfm,pan2024glomap, theia-manual} rely on an initial Structure-from-Motion step to recover a sparse reconstruction and camera poses, which are then input to a Multi-View Stereo (MVS) step to predict dense depth maps. Early work~\cite{schoenberger2016mvs} solved the task by carefully designing algorithms to match regions between images and optimizing for photometric and geometric consistency. 

To improve robustness to noisy real-world data, further works~\cite{xu2023unifying,demon,yao2018mvsnet} have focused on introducing learned modules alongside proven traditional matching algorithms, \eg, via image warping or cost volumes. UniMatch~\cite{xu2023unifying} extracts ResNet~\cite{he2016deep} features from the input images and uses a plane-sweeping method to match them; MVSNet~\cite{yao2018mvsnet} uses homography-warping to build a cost volume from which depth is regressed with 3D convolutions. However, these methods run into the same issues as their non-learnable counterparts: bad predictions in unmatched regions and reliance on precise camera poses.

Recent work has increasingly removed inductive biases and instead relied on discriminative features such as DINOv2~\cite{oquab2023dinov2}. DUSt3R~\cite{dust3r_cvpr24} first proved that plain neural networks are capable of predicting point maps without geometric processing, but it still required post-processing steps to retrieve camera poses and depth maps. Further works~\cite{wang2025vggt,keetha2026mapanything,depthanything3} have completely removed inductive biases and expensive bundle adjustment steps, only enforcing cross-attention~\cite{transformer} between input images. Nevertheless, their need for internet-scale training datasets to achieve robustness makes these models prohibitively expensive to replicate compared to inductive-bias models. In this paper, we strike a balance by using a model which uses monocular depth priors rooted in DINOv2's discriminative features and learns correspondences with a plane-sweeping method.

\noindent
\textbf{Single-Scene Novel View Synthesis.}
\label{subsubsec:single-scene-nvs}
Given a set of input images, NVS aims to generate images from new viewpoints. Since this task is fundamentally about recovering a dense scene reconstruction, it has typically shared the same inputs (posed images) and framework (per-scene optimization) as MVS. Similar to depth estimation, learning-based methods~\cite{mildenhall2020nerf, kerbl3Dgaussians, ye2025gsplat} dominate the field through differentiable representations and inverse-rendering techniques. 

To render images, most methods use volumetric rendering~\cite{levoy_volume,kajiya_volumetric} to integrate color along view-rays from a differentiable representation in a forward pass, and then update the representation to match the input images during a backward pass. NeRF~\cite{mildenhall2020nerf} uses MLPs to encode the scene as a continuous function, which models volume density and emitted radiance given a 3D point and viewing angle. While effective, this formulation requires repeated sampling of the neural implicit representation, making rendering and training slow.

3D Gaussian Splatting (3DGS)~\cite{kerbl3Dgaussians} instead models the scene with an explicit Gaussian point cloud. In doing so, color rendering is simplified to accumulating the contributions of intersected Gaussians. Further works~\cite{Huang2DGS2024,Yu2024GOF,Zhang2026GeometryGrounded} have focused on enforcing geometric constraints to achieve accurate geometry. While both 3DGS and NeRF achieve excellent visual fidelity, they require dense input views and per-scene training, preventing real-time reconstruction. In this paper, we use a feed-forward model to forgo time-consuming per-scene optimization and address the sparse-view constraint for geometric accuracy by introducing an edge-aware normal consistency loss. 

\noindent
\textbf{Feed-Forward Novel View Synthesis.}
\label{subsubsec:ffwd-nvs}
Motivated by advances in Multi-View Depth Estimation and Single-Scene Novel View Synthesis, recent works have combined these tasks to enable novel
view synthesis in a single forward pass. pixelSplat~\cite{charatan23pixelsplat} pioneered this approach by designing an epipolar transformer to match features between images and predict a depth distribution along view-rays. DepthSplat~\cite{xu2024depthsplat} relies instead on a cost volume formulation based on UniMatch's model and incorporates monocular depth features from Depth Anything V2~\cite{yang2024depth}. SurfSplat~\cite{he2026surfsplat} enforces a surface continuity prior to improve geometric accuracy.

Mirroring methods devoid of inductive biases in Multi-View Depth Estimation, recent works~\cite{jin2025lvsm,asim26scenetok,mitchel2026true} rely on transformers to learn to generate novel views. In doing so, they also forgo intermediate representations such as 3D Gaussian primitives. Relevant to our goal of geometric accuracy, these models complicate geometric self-supervision since there is no explicit structure to evaluate, and their latent features entangle geometric, visual, and camera pose attributes. 

Instead, we rely on the feed-forward Gaussian splatting model ReSplat~\cite{xu2025resplat} given its $16\times$ reduction in primitive count compared to per-pixel methods, which eases downstream tasks such as unbounded mesh extraction. Furthermore, we propose using a 3D smoothing filter to enable robust fine-tuning across different losses and decoders, allowing knowledge transfer from photometric quality to geometric accuracy.

\section{Preliminaries}
\label{sec:preliminaries}
In \Cref{subsec:gaussian-splatting}, we review the 3D Gaussian Splatting (3DGS)~\cite{kerbl3Dgaussians} rasterization process, along with extensions aimed at rendering normals and depth. We then discuss the sampling theorem in \Cref{subsec:sampling-theorem} and review solutions for degenerate cases where predicted Gaussians project to subpixel areas.

\subsection{3D Gaussian Splatting}
\label{subsec:gaussian-splatting}
Explicit differentiable representations are dominated by sets of 3D Gaussians~\cite{kerbl3Dgaussians} given their smooth, continuous, and anisotropic extent, which produces smooth gradients and enhances expressivity. Their appearance is dictated by opacity, $\alpha$, and spherical harmonics, which model view-dependent color, $\mathbf{c}$. Each primitive is defined, similarly to a Gaussian distribution, by its mean $\mathbf{p} \in \mathbb{R}^{3 \times 1}$ and covariance matrix $\mathbf{\Sigma} \in \mathbb{R}^{3\times3}$. To ensure the optimized covariance matrices are positive semi-definite, intermediate scale and quaternion rotations are used as parametrizations.

During rendering, given a camera pose, the 3D Gaussians are transformed to camera coordinates and projected to view-space to obtain 2D Gaussians, $\mathcal{G}^{2D}$. Then, colors are accumulated along a view-ray using alpha-blending. For intersected Gaussians $1, ..., K$, the volume rendering equation~\cite{kajiya_volumetric} is approximated as:
\begin{equation}
\label{eq:color-vre}
\bc(\bx) = \sum^K_{k=1} \bc_k\,\alpha_k\,\cG^{2D}_k(\bx) \prod_{j=1}^{k-1} (1 - \alpha_j\,\cG^{2D}_j(\bx)).
\end{equation}

\noindent
\textbf{Normals and Depth Rendering.}
Rendering accurate and consistent normal and depth maps is crucial for producing geometrically accurate Gaussian splatting scenes, allowing for evaluation against ground-truth recordings and geometric self-supervision. Follow-up methods~\cite{Huang2DGS2024, Yu2024GOF, Zhang2026GeometryGrounded} to 3D Gaussian Splatting have enabled normal rendering by using the volumetric rendering approximation from \Cref{eq:color-vre} and defining $\mathbf{n}_k$ as the normal vector of the $k$-th Gaussian along a ray. This yields the rendered normal:
\begin{equation}
\label{eq:normal-vre}
\mathbf{n}(\bx) = \sum^K_{k=1} \mathbf{n}_k\,\alpha_k\,\cG^{2D}_k(\bx) \prod_{j=1}^{k-1} (1 - \alpha_j\,\cG^{2D}_j(\bx)).
\end{equation}

However, following the same strategy for depth fails to produce consistent depth renderings, as neighboring pixel-rays may not agree on a common surface and are susceptible to outlier Gaussians. Furthermore, the approximation in \Cref{eq:normal-vre} is fundamentally flawed for depth renderings since the 3D to 2D Gaussian projection removes depth information.

Gaussian Opacity Fields (GOF)~\cite{Yu2024GOF} and Geometry-Grounded Gaussian Splatting (GGGS)~\cite{Zhang2026GeometryGrounded} solve this issue by projecting the intersected 3D primitives along view-rays to get 1D Gaussians, which preserve depth information. In doing so, they define a smooth, continuous transmittance (the product term inside \Cref{eq:normal-vre}) instead of a staircase function. Next, they treat the 0.5 transmittance level-set as the environment's surface. Since transmittance along a ray is the probability that the ray has not yet been absorbed, it is a monotonically decreasing function that can be searched iteratively for the 0.5-value crossing to render consistent and accurate depth maps. During depth rendering, GGGS narrows down on the surface level-set, in contrast to GOF, which computes the Gaussian maximum evaluation along a ray, producing yet again a staircase transmittance function and, ultimately, inconsistent and noisy depth maps.

In practice, GGGS only uses the 1D Gaussian formulation during depth rendering and reverts to projecting the primitives to 2D Gaussians during normal and color rendering. GOF completely avoids 2D Gaussians for accumulating normal and color values during rendering. Lastly, both GOF and GGGS extract meshes through Marching Tetrahedra~\cite{marching_tetrahedra} on an adaptive tetrahedral grid, which allows for unbounded scene mesh extraction, matching our stated goal.

\subsection{Sampling Theorem for Gaussian Splatting}
\label{subsec:sampling-theorem}
The Nyquist-Shannon Sampling Theorem~\cite{nyquist,nyquist_shannon} from signal processing establishes the following requirement needed to reconstruct a continuous signal from discrete samples:

\begin{nyquist}
The continuous signal must be band-limited with bandwidth $\nu$, and the sampling rate $\hat\nu$ must be at least twice the highest frequency present in the continuous signal: $\hat\nu\geq2\nu$.
\end{nyquist}
In Gaussian splatting, ignoring the Nyquist Conditions leads to aliasing issues when decreasing the sampling rate, and high frequency artifacts when increasing it, due to thin, degenerate Gaussians. For optimization routines, Mip-Splatting~\cite{Yu2024MipSplatting} proposes a 2D Mip filter applied on the projected 2D Gaussians to solve dilation and aliasing issues, and a 3D smoothing filter defined by the sampling rate of all training views to solve Gaussian erosion. In this paper, we identify that the 3D smoothing filter is essential for stable encoder training and fine-tuning with new decoders and losses.

\section{Method}
\label{sec:method}
\begin{figure}[tb]
  \centering
  \includegraphics[width=0.9\textwidth]{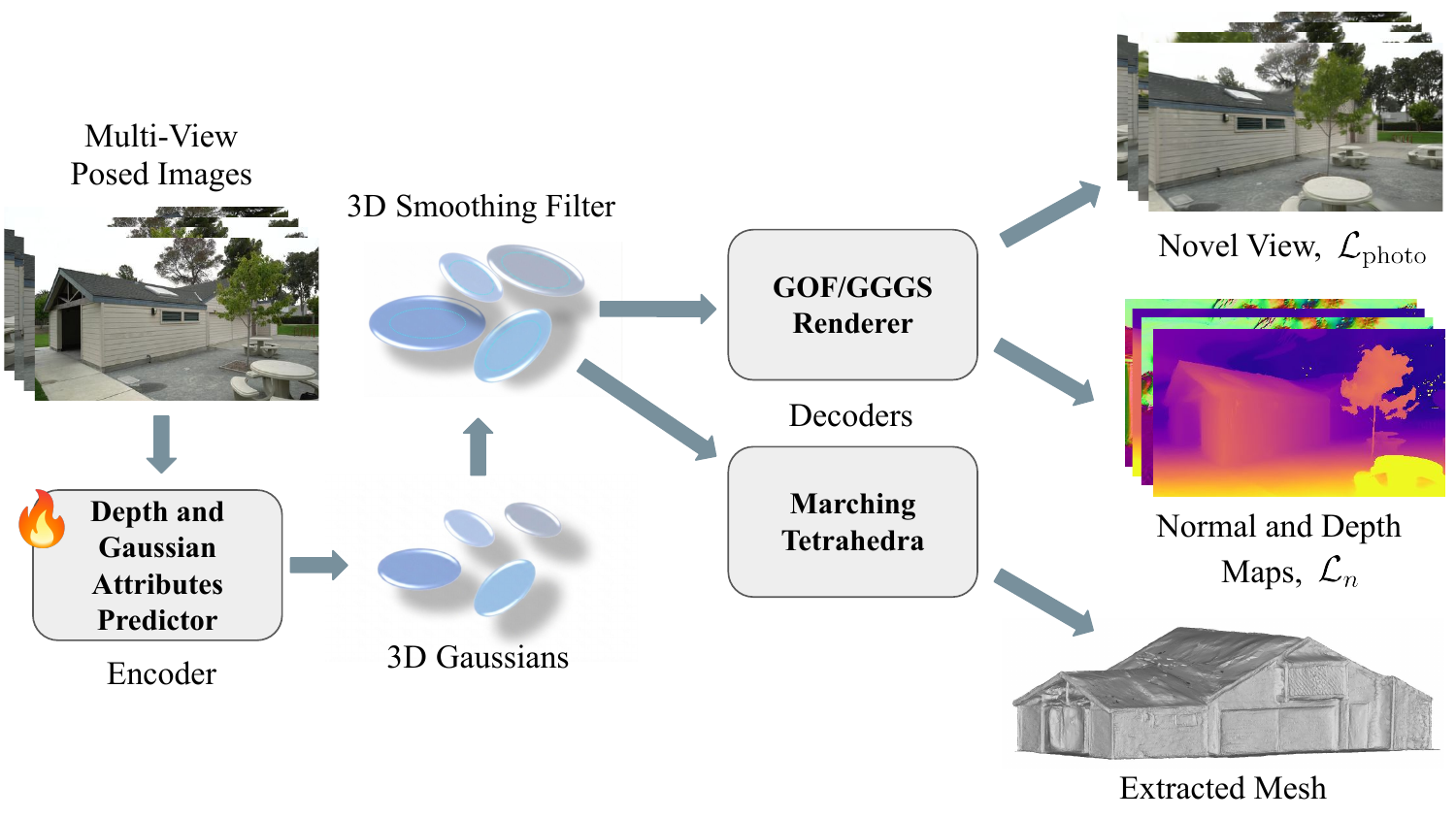}
  \caption{\textbf{Our framework}. Given $N$ multi-view posed images, our encoder outputs $M=N \times\frac{HW}{16}$ Gaussian primitives reconstructing the scene. Next, a 3D smoothing filter ensures that the predicted Gaussians do not exceed the Nyquist frequency. During training, novel color, depth, and normal maps are rendered with the GOF/GGGS renderer and are used to train the encoder with photometric and geometric losses. At testing time, the decoder can be switched to a Marching Tetrahedra algorithm, capable of handling unbounded scenes, to extract geometrically accurate meshes.
  }
  \label{fig:framework}
\end{figure}

We present a feed-forward framework for predicting geometrically accurate 3D Gaussians. Our framework comprises a learnable encoder (\Cref{subsec:encoder}) that generates these primitives, and dedicated decoders (\Cref{subsec:gaussian-splatting}) for efficient image rendering and mesh extraction. An overview of the framework is shown in \cref{fig:framework}.

To achieve high geometric fidelity, we introduce an edge-aware normal consistency loss tailored specifically for sparse Gaussian representations (\Cref{subsec:edge-aware-nc-loss}). Furthermore, we propose a 3D smoothing filter to ensure robust fine-tuning of the encoder across different losses and decoders (\Cref{subsec:3d-filter}, \Cref{subsec:robust-encoder-finetuning}). Together, these additions allow us to take an encoder previously trained for visual fidelity and fine-tune it for superior geometric accuracy. 

\subsection{Encoder}
\label{subsec:encoder}
Our encoder is based on the ReSplat~\cite{xu2025resplat} architecture which predicts $16 \times$ fewer Gaussians than per-pixel approaches~\cite{charatan23pixelsplat,chen2024mvsplat,xu2024depthsplat,he2026surfsplat}. Given $N$ input images $\{{\bm I}^{i}\}_{i=1}^N ({\bm I}^i \in \mathbb{R}^{H \times W \times 3}$) with their intrinsic $\{ \bm {K}^i \}_{i=1}^N$ ($\bm {K}^i \in \mathbb{R}^{3 \times 3}$) and extrinsic $\{ ({\bm R}_i, {\bm t}_i) \}_{i=1}^N$ (${\bm R}_i \in \text{SO}(3), {\bm t}_i \in \mathbb{R}^3 $) matrices, the encoder predicts a set of 3D Gaussian primitives~\cite{kerbl3Dgaussians} $\mathcal{G}$ with attributes $\{ (\mathbf{p}_k, \alpha_k, \mathbf{\Sigma}_k, \bm{\mathrm{sh}}_k ) \}_{k=1}^M$ to model the scene, where $M=N \times \frac{HW}{16}$ is the total number of Gaussian primitives. 

Following previous works~\cite{charatan23pixelsplat,chen2024mvsplat,xu2024depthsplat}, the encoder has a depth prediction module to predict the Gaussians' means and a lightweight network to predict other remaining attributes. The depth prediction network consists of a multi-view branch which extracts ResNet~\cite{he2016deep} features from the input images and refines them with self- and cross-attention~\cite{transformer} layers. Next, cost volumes are constructed for each view with a plane-sweeping algorithm to match features across views and depths. Since plane-sweeping methods fail in texture-less regions and unmatched regions such as occlusions, a single-view branch is used to extract monocular depth features from Depth Anything V2~\cite{yang2024depth}. These features are concatenated and fused into refined cost volumes from which depth maps are predicted. To achieve a $16 \times$ reduction in primitives, the depth predictions are resized to 1/4 of the original resolution. To compensate for the reduced primitive count, the features are unprojected to 3D space using the camera poses and predicted depths and refined with six alternating blocks of kNN attention and low-resolution global attention. Finally, an MLP takes in the refined features and predicts the remaining Gaussian attributes.

\subsection{Edge-Aware Normal Consistency Loss}
\label{subsec:edge-aware-nc-loss}
To ensure that the 3D Gaussians are aligned with the environment's surfaces we use a variation of the normal consistency loss introduced by 2DGS~\cite{Huang2DGS2024}:
\begin{equation}
\label{loss:eanc}
\mathcal{L}_n = \bS \odot (1-\bn^\mathrm{T}\bN),
\end{equation}
where $\bn$ is the rendered normal map defined in \Cref{eq:normal-vre}, $\bN$ is the estimated normal surface as computed by the finite-difference gradient of the depth map, and $\bS$ is the image smoothness map computed from the ground-truth color image:
\begin{equation}
\label{eq:smooth_map}
\bS = e^{-|\partial_x \bm{I}|} + e^{-|\partial_y \bm{I}|}.
\end{equation}
The second term of $\mathcal{L}_{n}$ encourages the primitives along a ray to have normals which match the estimated normal surface and vice versa. As for the first term, we observe that in the optimization case, densification routines allocate Gaussians in high-frequency image regions. However, for our feed-forward case, the Gaussians are allocated equally across the scene making the original normal consistency loss susceptible to under-reconstructed high-frequency regions in rendered normal maps. Since low-gradient regions in images typically belong to smooth surfaces, the edge-aware normal consistency loss focuses on these as determined by the image smoothness map. 

\subsection{3D Smoothing Filter for Feed-Forward Gaussian Splatting}
\label{subsec:3d-filter}
As discussed in \Cref{subsec:gaussian-splatting}, renderers dedicated to geometric accuracy, such as GOF~\cite{Yu2024GOF}, do not project the 3D Gaussians to view-space during color rendering and instead re-parametrize the primitive to be a 1D Gaussian along the view-ray. We find that this approach is particularly susceptible to severe aliasing artifacts as the 2D Mip filter proposed by Mip-Splatting~\cite{Yu2024MipSplatting} is specifically designed to widen the projected 2D Gaussians' view-space footprint. Furthermore, while in GOF's optimization case, learning can recover from aliasing artifacts as shown by their success in per-scene training, we find that for feed-forward networks, degenerate cases where the 3D Gaussians project to subpixel regions produce large gradients leading to model divergence and unstable training.

Thus, to successfully avoid model divergence and ensure robust training, we unproject the 2D Mip filter to 3D space given depth $d$ and camera focal length $f$. Following linear perspective and the pinhole camera model:
\begin{equation}
\label{eq:unproject-2d-mip-filter}
\underset{\text{2D Mip filter}}{\mathcal{N}(0, s\bI) \vphantom{\frac{d}{f}} \rule[-1.5em]{0pt}{0pt}} 
\xrightarrow{\text{unproject}} 
\underset{\text{3D Smoothing Filter}}{\mathcal{N}\left(0, s\frac{d}{f}\bI\right). \rule[-1.5em]{0pt}{0pt}}
\end{equation}

Since the scale of the 3D filter is then $s/\nu$, where $\nu=f/d$ is the sampling frequency, we observe that the 2D and 3D filters from Mip-Splatting are closely related, with a few important distinctions. Mip-Splatting's 3D filter uses the maximum frequency across all cameras because primitives receive direct updates from all views during optimization routines.
In our feed-forward case, each Gaussian $k$ is generated by, and anchored to, a specific source view $i$. By tying the primitive to its source view, we can streamline the filter by making the following assumptions:
\begin{enumerate}
\item 
\textbf{Maximum sampling rate:} To satisfy the Nyquist-Shannon Sampling Theorem (\cref{subsec:sampling-theorem}), the spatial frequency of the predicted primitives must be band-limited to prevent aliasing. Because Gaussian $k$ is extracted from the discrete pixel grid of its source view $i$, its maximum sampling rate is inherently bounded by that view's focal length $f_i$ and the primitive's predicted depth $d_k$. Although cross-attention layers exist, primitives do not receive direct geometric updates from all views like they do in per-scene optimization.

\item
\textbf{Handling degenerate projections:} Our training and test datasets feature inward-facing circular camera trajectories. In this setup, a primitive's minimum sampling rate is also effectively governed by its source view's depth $d_k$ and focal length $f_i$. This ensures our filter reliably resolves degenerate subpixel projections for input and target views without requiring 2D filters at render-time, which is crucial for the GOF decoder.
\end{enumerate}

Therefore, we can completely avoid the expensive computation of the maximum frequency across all cameras. Instead, we efficiently compute the 3D smoothing filter for the $k$-th Gaussian using the predicted depth $d_k$ and focal length $f_i$ of its source view:

\begin{equation}
\label{eq:3d-filter}
\mathcal{N}\left(0, s\frac{d_k}{f_i}\bI\right) = \mathcal{N}\left(0, \frac{s}{\nu_k}\bI\right).
\end{equation}

\subsection{Robust Encoder Fine-Tuning}
\label{subsec:robust-encoder-finetuning}
As models and training datasets become increasingly large, we focus on enabling robust fine-tuning to reduce training times while achieving the highest possible geometric and visual fidelity. Our core hypothesis is that a model excelling in visual quality can also recover accurate geometry when properly constrained. To test this, we start with a ReSplat encoder that was previously trained with the gsplat~\cite{ye2025gsplat} decoder (which lacks normal map rendering), utilizing solely photometric losses, and without the 3D smoothing filter. We then fine-tune this encoder using the geometry-focused GOF~\cite{Yu2024GOF} and GGGS~\cite{Zhang2026GeometryGrounded} decoders, enabling the filter, and supervising it with both photometric and geometric consistency losses (\cref{loss:eanc}).

\begin{figure}[tb]
  \centering
  \includegraphics[width=0.7\textwidth]{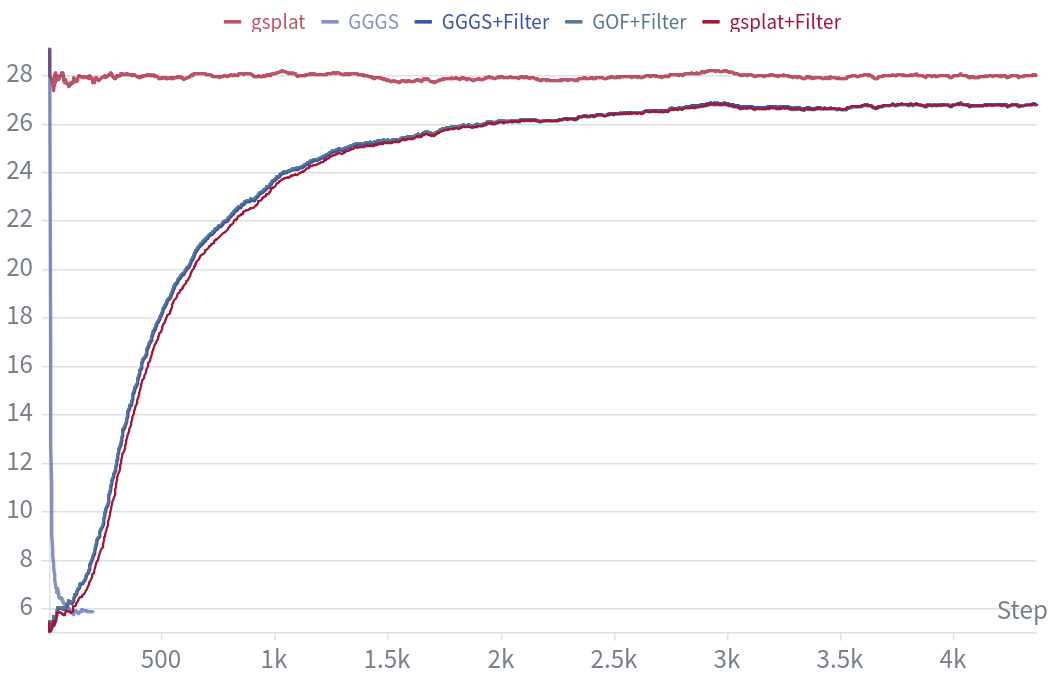}
  \caption{\textbf{Fine-Tuning ReSplat encoder trained with gsplat}. We report the PSNR metric over 4k fine-tuning steps across different decoder configurations. While the GGGS decoder immediately diverges after the first step due to backward pass instabilities, enabling our proposed 3D smoothing filter allows the encoder to stabilize and recover. After 4k steps, all configurations utilizing the filter converge to similar visual quality, scoring within 1.6dB of the gsplat baseline.
  }
  \label{fig:robust_finetuning}
\end{figure}

In theory, switching the gsplat and GGGS decoders should work out of the box as both share the rendering forward pass including projecting the primitives to 2D view-space and applying the 2D Mip filter from Mip-Splatting. However, after the very first fine-tuning step, the encoder output diverges. Given our previous experience identifying degenerate subpixel projections as the main cause for unstable training with the GOF decoder, we inspect the 2D Mip filter backward pass. We find that in gsplat, the chain rule math has been simplified avoiding divisions that lead to large gradients. On the other hand, in the vanilla Gaussian splatting codebase from which GOF and GGGS build upon, the derivatives are computed through intermediate variables and divisions by zero are avoided by completely stopping gradient flow. We propose enabling the 3D smoothing filter from \Cref{subsec:3d-filter} even if the loaded encoder was not trained with it. In doing so, we ensure that the backward pass of both decoders operates in the same domain forgoing degenerate inputs for which their behavior diverges. In \Cref{fig:robust_finetuning}, we see the successful recovery when enabling the 3D smoothing filter while fine-tuning a gsplat-trained encoder with different decoders.

\subsection{Training Loss}
\label{subsec:training_loss}
Our model is trained with photometric and geometric losses on novel views. Following prior works~\cite{xu2025resplat}, the photometric loss, $\mathcal{L}_\text{photo}$, is a linear combination of $\ell_1$ and a perceptual loss, $\ell_\text{perceptual}$, with loss weights of 1 and 0.5, respectively. The geometric loss $\mathcal{L}_n$ is the edge-aware normal consistency loss introduced in \Cref{subsec:edge-aware-nc-loss} with loss weight of 0.02.

\section{Experiments}
\label{sec:experiments}

\subsection{Implementation}
\label{subsec:implementation}
We implement our method in PyTorch~\cite{pytorch}. For the kNN attention module, 16 neighbors are used following Point Transformer~\cite{pttransformer}. For the 3D smoothing filter, $s=0.1$, matching the 2D Mip filter in Mip-Splatting. We train our model with 8 input views at 256 $\times$ 448 resolution on the DL3DV dataset. Fully trained models are initialized with a pre-trained DepthSplat~\cite{xu2024depthsplat} depth model while fine-tuned models are initialized with ReSplat view synthesis models and are trained with half the number of steps as fully trained models.

\subsection{Model Variants}
\label{subsec:model-variants}
To save compute when evaluating the performance of fine-tuned or fully trained models with the normal consistency loss and its edge-aware variant, we use a lighter ViT-S backbone. To identify the different models, we abbreviate RS for ReSplat, EANC for edge-aware normal consistency loss, and NC for normal consistency loss without the image smoothness weighing. Later on, when benchmarking against other methods we use a ViT-B backbone.

\subsection{Evaluation}
\begin{table}[tb]
\centering
\small
\setlength{\tabcolsep}{4pt}
\caption{\textbf{Comparison of Training Strategies (Fully Trained \vs Fine-Tuned) and Normal Consistency Variants.} We evaluate encoders trained with photometric and geometric losses to test the performance of different training strategies and geometric losses. Our results indicate that fine-tuning the model with the edge-aware normal consistency loss achieves best mesh metrics. Best and second best results are \textbf{bolded} and \underline{underlined}.}
\label{tab:fully_fine_eanc}
\resizebox{\linewidth}{!}{
\begin{tabular}{ll cccc c ccc}
\toprule
\textbf{Encoder} & & \multicolumn{4}{c}{\textbf{Depth}} & \multicolumn{1}{c}{\textbf{Mesh}} & \multicolumn{3}{c}{\textbf{Photo}} \\
\cmidrule(lr){3-6} \cmidrule(lr){7-7} \cmidrule(lr){8-10}
\textbf{Decoder} & \textbf{Configuration} & AbsRel $\downarrow$ & RMSE $\downarrow$ & MAE $\downarrow$ & $\delta_1 \uparrow$ & F1-score $\uparrow$ & PSNR $\uparrow$ & SSIM $\uparrow$ & LPIPS $\downarrow$ \\
\midrule
\textbf{RS} & NC Full & \underline{0.091} & 4.595 & 38.673 & \underline{0.900} & 0.108 & 21.11 & 0.619 & 0.314 \\
\textbf{GOF} & NC Fine-tune & \textbf{0.084} & \underline{4.444} & \underline{37.249} & \textbf{0.907} & \underline{0.115} & \textbf{21.67} & \underline{0.645} & \underline{0.291} \\
 & EANC Fine-tune & \textbf{0.084} & \textbf{4.429} & \textbf{35.871} & \textbf{0.907} & \textbf{0.119} & \underline{21.63} & \textbf{0.646} & \textbf{0.290} \\
\midrule
\textbf{RS} & NC Full & 0.093 & 4.661 & \underline{36.023} & 0.897 & 0.100 & 21.02 & 0.615 & 0.322 \\
\textbf{GGGS} & NC Fine-tune & \underline{0.085} & \textbf{4.414} & 36.415 & \textbf{0.909} & \underline{0.105} & \textbf{21.81} & \textbf{0.652} & \textbf{0.288} \\
 & EANC Fine-tune & \textbf{0.084} & \underline{4.471} & \textbf{35.182} & \underline{0.906} & \textbf{0.108} & \underline{21.72} & \underline{0.647} & \underline{0.293} \\
\bottomrule
\end{tabular}
}
\end{table}

\label{subsec:evaluation}
We focus on outdoor unbounded scenarios and use 10 outdoor scenes from the DL3DV~\cite{ling2024dl3dv} test set, outdoor scenes from Mip-NeRF 360~\cite{barron2022mipnerf360}, and the smaller scenes from the Tanks and Temples (TnT)~\cite{Knapitsch2017} dataset (no Courthouse nor Meetingroom). To get pseudo ground-truth depth, we optimize each scene for 30k steps on the full view set using the state-of-the-art GGGS~\cite{Zhang2026GeometryGrounded}. We select input views in a similar circular trajectory as the ones in DL3DV and average results for 8 and 16 input views. When reporting mesh metrics, ground-truth point clouds from TnT are used.

\subsubsection{Fine-Tuned Encoder's Performance.}
\label{subsubsec:full-ft-performance}
In \Cref{tab:fully_fine_eanc}, we compare encoders trained with the 3D smoothing filter and photometric and normal consistency losses. The first one, NC Full, is fully trained and initialized from a checkpoint trained for depth estimation following the training strategy in ReSplat. The second one, NC Fine-tune, is initialized from a checkpoint trained for visual quality and using gsplat as its decoder but without the 3D smoothing filter. Note that the fine-tuned encoder performs best across most depth, mesh, and visual quality metrics. Crucially, our fine-tuning strategy achieves superior performance while significantly reducing computational overhead. For the fine-tuned encoders, using the GOF decoder produces better mesh accuracy but worse visual quality than using the GGGS decoder. We attribute GOF's better mesh accuracy to its volumetric color rendering along each ray, whereas GGGS shares gsplat's view-space color rasterization, explaining its better visual quality.

\subsubsection{Edge-Aware Normal Consistency Ablation.}
\label{subsubsec:eanc-ablation}
\begin{figure*}[t]
    \centering
    \setlength{\tabcolsep}{0.1em}
    \renewcommand{\arraystretch}{0.4}
    
    \newcommand{\ctxwidth}{0.12\textwidth} %
    
    \vspace{0.5em} %
    
    \newcommand{\fivewidth}{0.196\textwidth} %
    
    \begin{tabular}{ccccc}
    \includegraphics[width=\fivewidth]{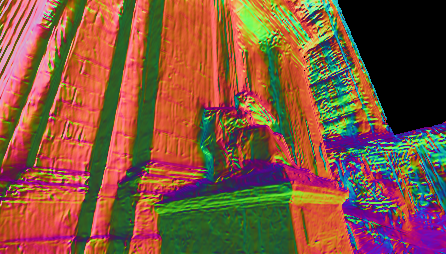}&
    \includegraphics[width=\fivewidth]{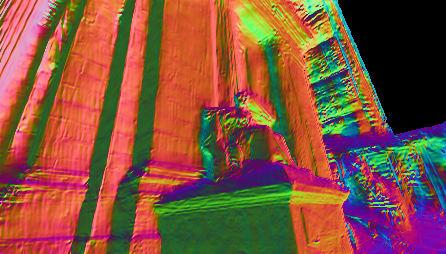}&
    \includegraphics[width=\fivewidth]{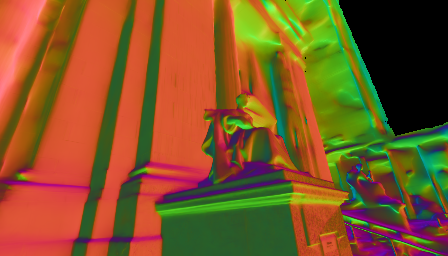}&
    \includegraphics[width=\fivewidth]{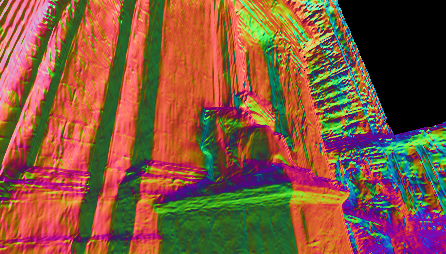}&
    \includegraphics[width=\fivewidth]{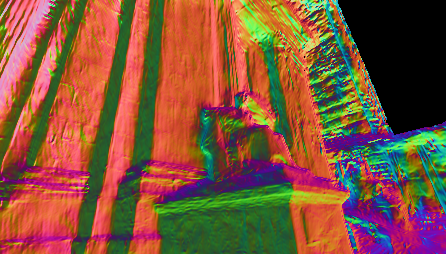}
    \\
    \begin{tabular}{@{}c@{}}NC \\ GOF\end{tabular} & 
    \begin{tabular}{@{}c@{}}EANC \\ GOF\end{tabular} & 
    \begin{tabular}{@{}c@{}}GT\end{tabular} & 
    \begin{tabular}{@{}c@{}}NC \\ GGGS\end{tabular} & 
    \begin{tabular}{@{}c@{}}EANC \\ GGGS\end{tabular} 
    \end{tabular}
    
    \caption{\textbf{Depth Normal Comparison on DL3DV test set.} Normal maps computed from the depth renderings of the predicted Gaussians. By training with the edge-aware normal consistency loss, the encoder learns to predict Gaussians which produce flat surfaces as shown in the building's walls on the left-side of the image.}
    \label{fig:depth_normal_cols_032_nosky}
\end{figure*}

We analyze the behavior of the edge-aware normal consistency loss in fine-tuned models and find that the edge-aware variant performs best in mesh metrics and the Mean Angular Error (MAE) metric of the normal map computed from the rendered depth maps. As evident in \Cref{fig:depth_normal_cols_032_nosky}, using the edge-aware normal consistency loss produces smoother surfaces along the building's façade as it guides the encoder to predict Gaussians that lie flat along surfaces. When it comes to depth and visual quality metrics, both variants score similarly and there is no consensus on which setup is better across the GOF and GGGS decoders. Results are shown in \Cref{tab:fully_fine_eanc}.

\begin{table}[tb]
\centering
\small
\setlength{\tabcolsep}{4pt}
\caption{\textbf{Depth, mesh, and method efficiency metrics.} We evaluate current state-of-the-art depth estimation and Gaussian splatting methods. Our method achieves the best mesh accuracy, F1-score, just behind per-scene optimization while being almost $200 \times$ faster in reconstruction and mesh extraction times. Best and second best results are \textbf{bolded} and \underline{underlined} respectively.}
\label{tab:geo_bench}
\resizebox{\linewidth}{!}{
\begin{tabular}{l ccccccccccc}
\toprule
\multirow{2}{*}{\textbf{Method}} & \multicolumn{4}{c}{\textbf{Depth}} & \multicolumn{4}{c}{\textbf{F1-score}} & \multicolumn{3}{c}{\textbf{Efficiency}} \\
\cmidrule(lr){2-5} \cmidrule(lr){6-9} \cmidrule(lr){10-12}
 & AbsRel $\downarrow$ & RMSE $\downarrow$ & MAE $\downarrow$ & $\delta_1 \uparrow$ & Barn & Caterpillar & Ignatius & Truck & Faces (M) $\downarrow$ & Recon. (s) $\downarrow$ & Extr. (s) $\downarrow$ \\
\midrule
SurfSplat & 0.210 & 131.629 & 60.414 & 0.783 & 0.058 & 0.046 & 0.095 & 0.095 & 9.41 & \underline{0.43} & 56.94 \\
DA3 & 0.119 & 6.360 & \textbf{30.568} & 0.833 & 0.008 & 0.016 & 0.023 & 0.043 & 21.07 & 4.67 & 79.50 \\
RS+gsplat & 0.087 & \underline{4.201} & 49.649 & \underline{0.917} & 0.081 & 0.064 & 0.141 & 0.134 & \textbf{1.81} & \textbf{0.32} & \underline{14.11} \\
RS+GGGS+EANC & \underline{0.081} & \textbf{4.133} & \underline{34.040} & \textbf{0.919} & 0.110 & \underline{0.079} & 0.173 & \underline{0.184} & \underline{1.89} & \textbf{0.32} & \textbf{1.45} \\
RS+GOF+EANC & \textbf{0.080} & 4.202 & 34.530 & \underline{0.917} & \underline{0.113} & \underline{0.079} & \underline{0.175} & 0.179 & 2.60 & \textbf{0.32} & 20.93 \\
GGGS Opt. 4k & 0.086 & 4.659 & 35.094 & 0.915 & \textbf{0.123} & \textbf{0.106} & \textbf{0.259} & \textbf{0.203} & 14.69 & 246.72 & 95.65 \\
\bottomrule
\end{tabular}
}
\end{table}

\subsubsection{Geometry Benchmark Comparisons.}
\label{subsubsec:benchmark-comparisons}
In this section we compare our best-performing configuration, edge-aware normal consistency loss and fine-tuned encoder, against current methods in feed-forward Gaussian splatting, depth estimation, and Gaussian splatting optimization routines. For feed-forward models, ours and SurfSplat, we use a ViT-B backbone. In \cref{tab:geo_bench} we compare methods across depth metrics, F1-score on the extracted meshes, and mesh compactness as well as reconstruction time and mesh extraction time. 

Focusing on the depth metrics, enabling the edge-aware normal consistency loss has little impact on depth accuracy but significantly improves the depth map's gradients by lowering the MAE metric. Other methods, SurfSplat~\cite{he2026surfsplat} and DA3~\cite{depthanything3}, which have been trained on datasets which prioritize indoor scenes fail badly at producing accurate depth maps. In the case of DA3, which we run with COLMAP camera poses, depth prediction consistently fails to place the surfaces at the correct depth as seen in \cref{fig:MeshCompBench} for the Barn and Truck scenes where objects are closer than they should be. 

Moving on to mesh metrics, our method performs best among feed-forward methods and comes close to per-scene optimization for 4k iterations. Furthermore, our method is substantially more efficient against per-scene optimization. Since our encoder predicts $16 \times $ fewer Gaussians than pixels, mesh extraction is faster as the tetrahedral grid adapts to the number of Gaussians in the scene. GGGS also introduces a faster transmittance computation further reducing the time compared to using GOF as decoder. Finally, our output mesh is lighter than SurfSplat's and DA3's, which rely on TSDF, a method to extract meshes from depth maps that is susceptible to artifacts in the sparse-view case.

Lastly, as seen in \cref{fig:MeshCompBench}, while the per-scene optimization meshes achieve better F1-score metrics, there are artifacts due to over-fitting to the input images. This is most clearly seen for the Barn scene where the façade of the building has many ridges instead of being flat. On the other hand, DA3 and our method, leverage learned priors, thus correctly predicting walls and floors to be flat.
\begin{figure*}[t]
    \centering
    \setlength{\tabcolsep}{0.1em}
    \renewcommand{\arraystretch}{0.4}
    
    \newcommand{\fivewidth}{0.195\textwidth} 
    
    \begin{tabular}{ccccc}
    \includegraphics[width=\fivewidth]{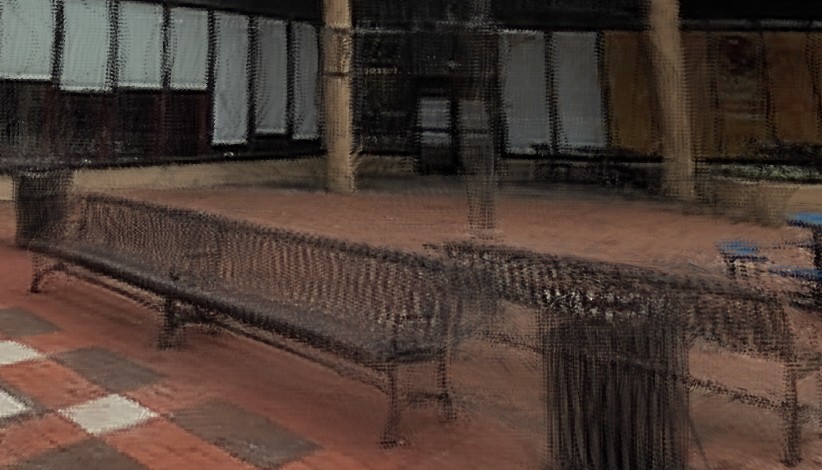}&
    \includegraphics[width=\fivewidth]{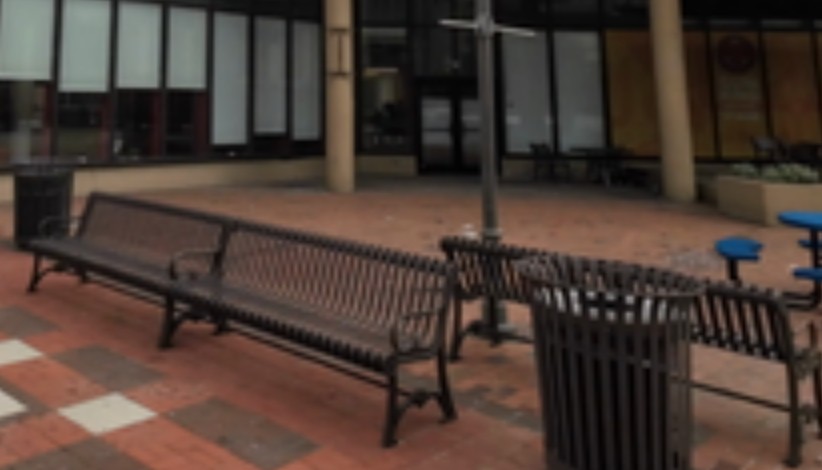}&
    \includegraphics[width=\fivewidth]{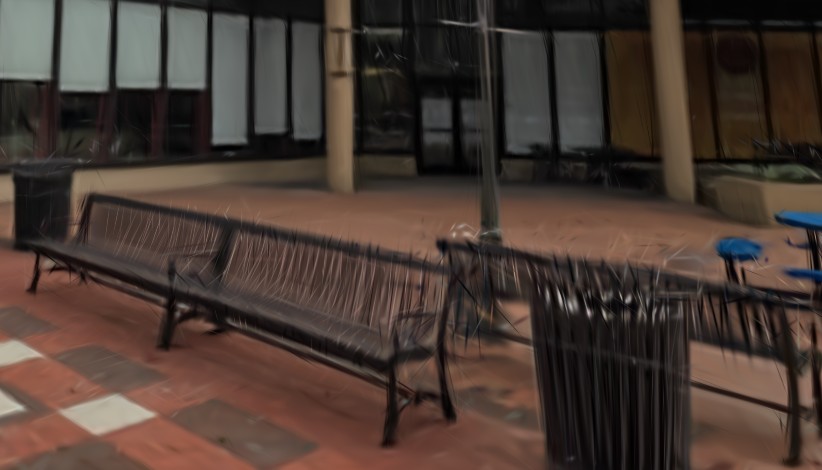}&
    \includegraphics[width=\fivewidth]{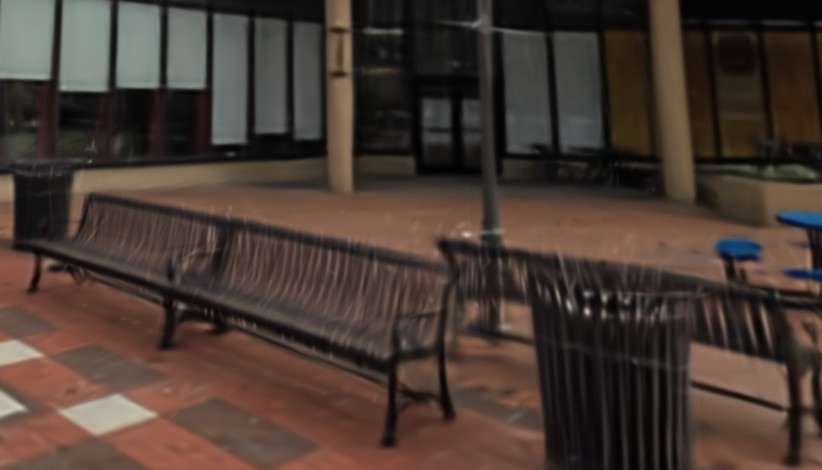}&
    \includegraphics[width=\fivewidth]{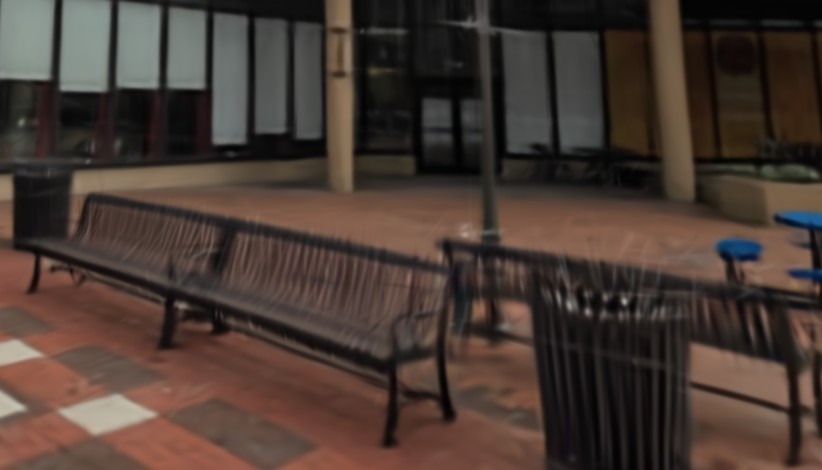}
    \\
    \includegraphics[width=\fivewidth]{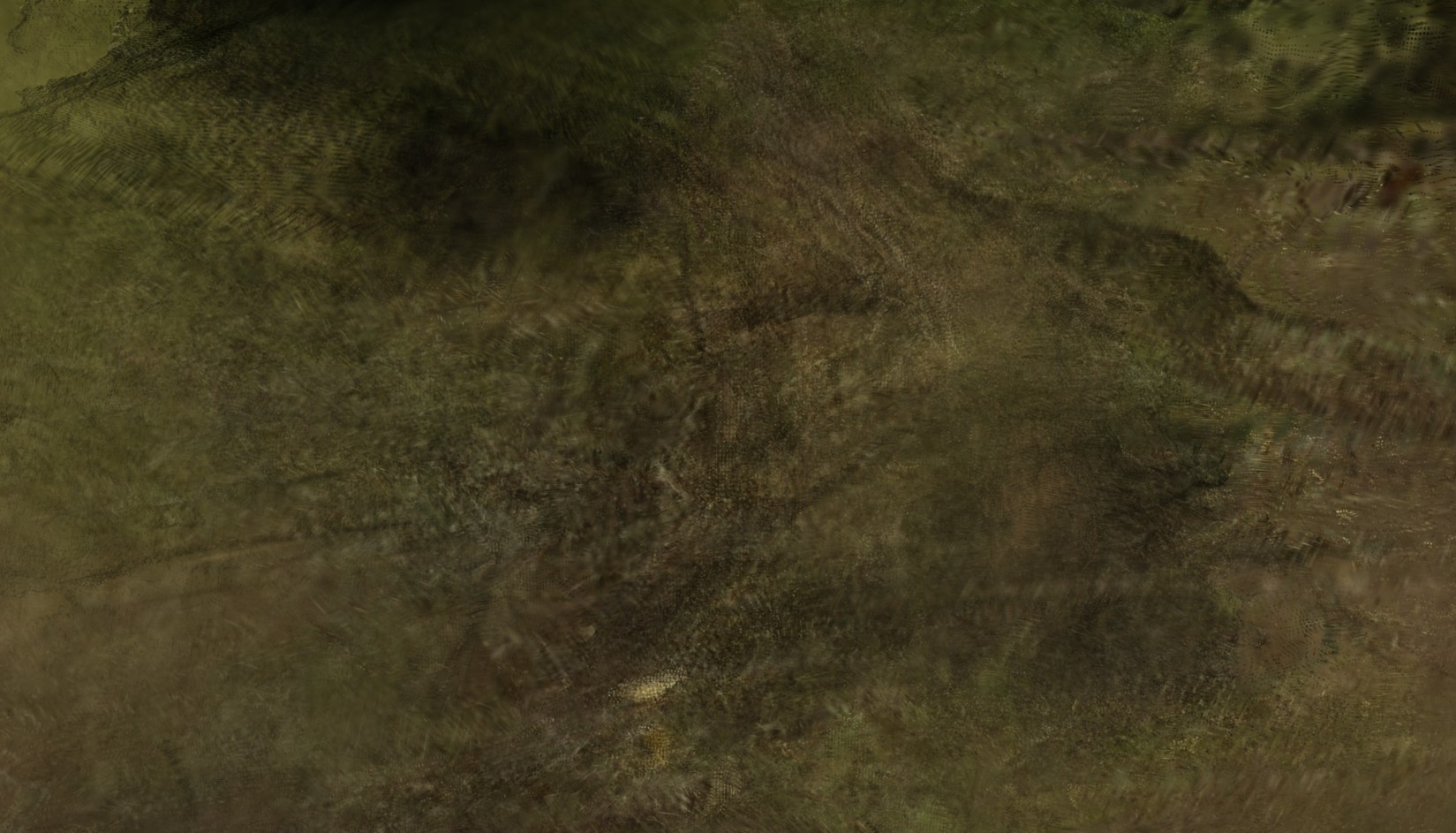}&
    \includegraphics[width=\fivewidth]{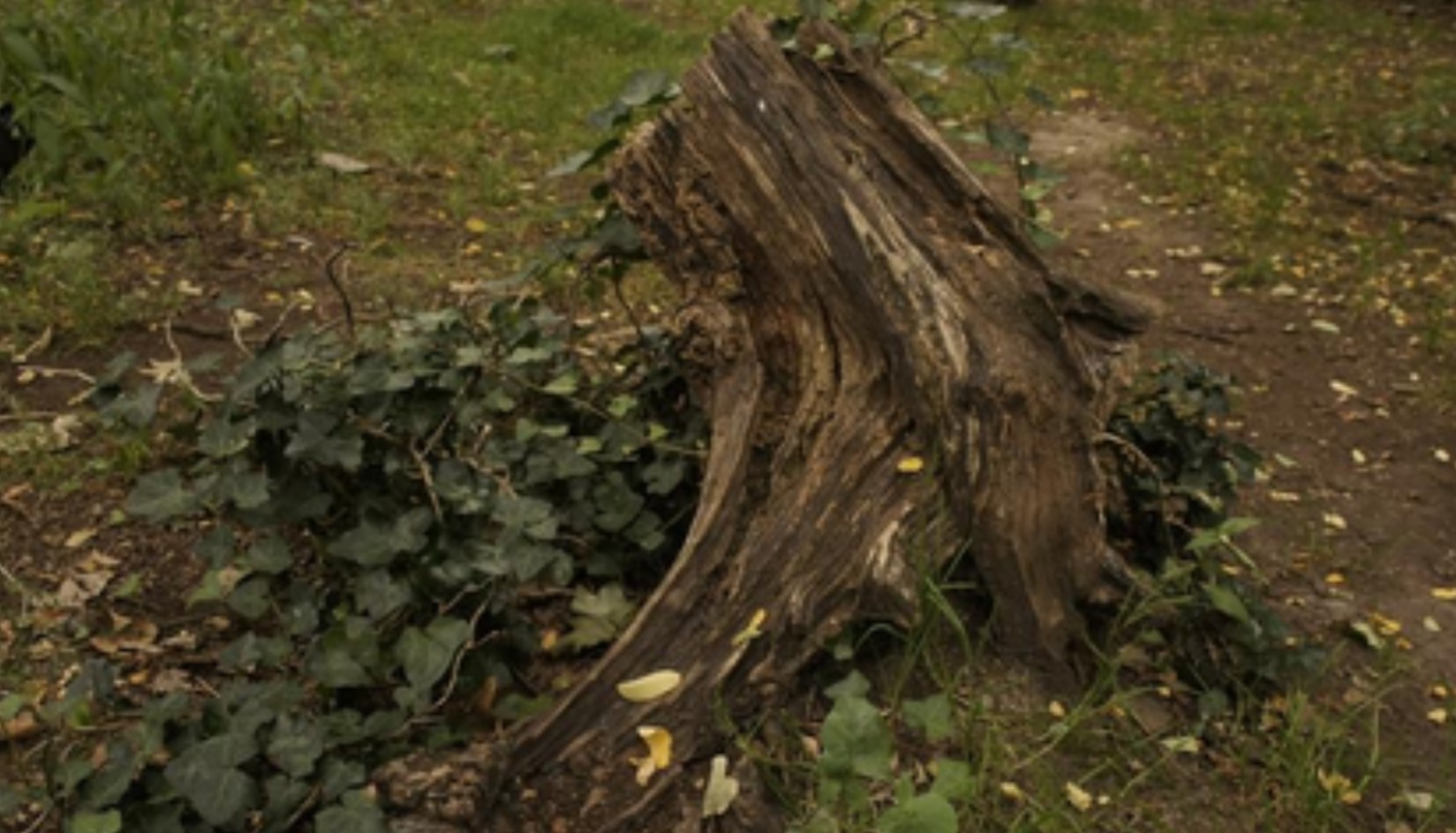}&
    \includegraphics[width=\fivewidth]{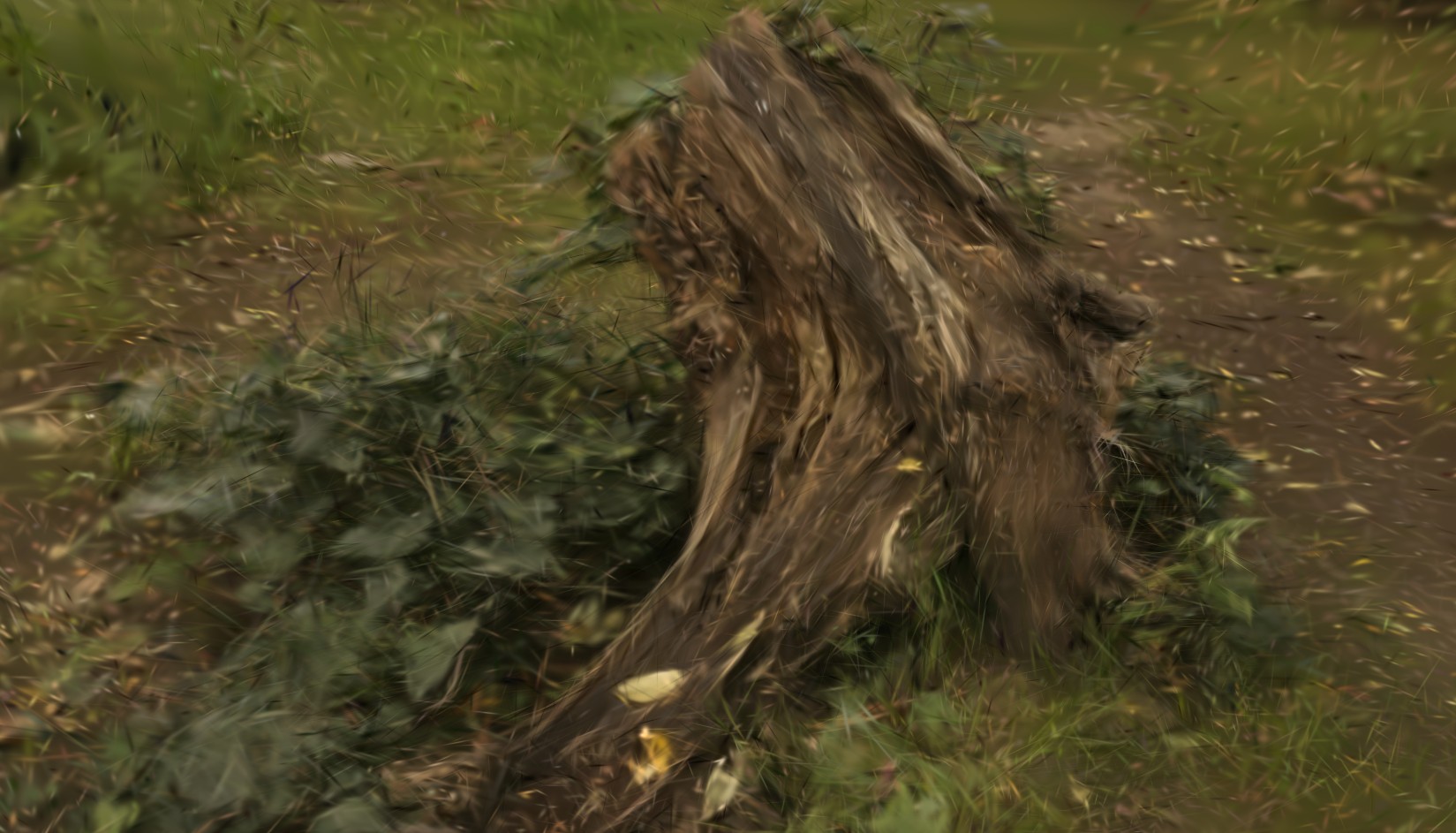}&
    \includegraphics[width=\fivewidth]{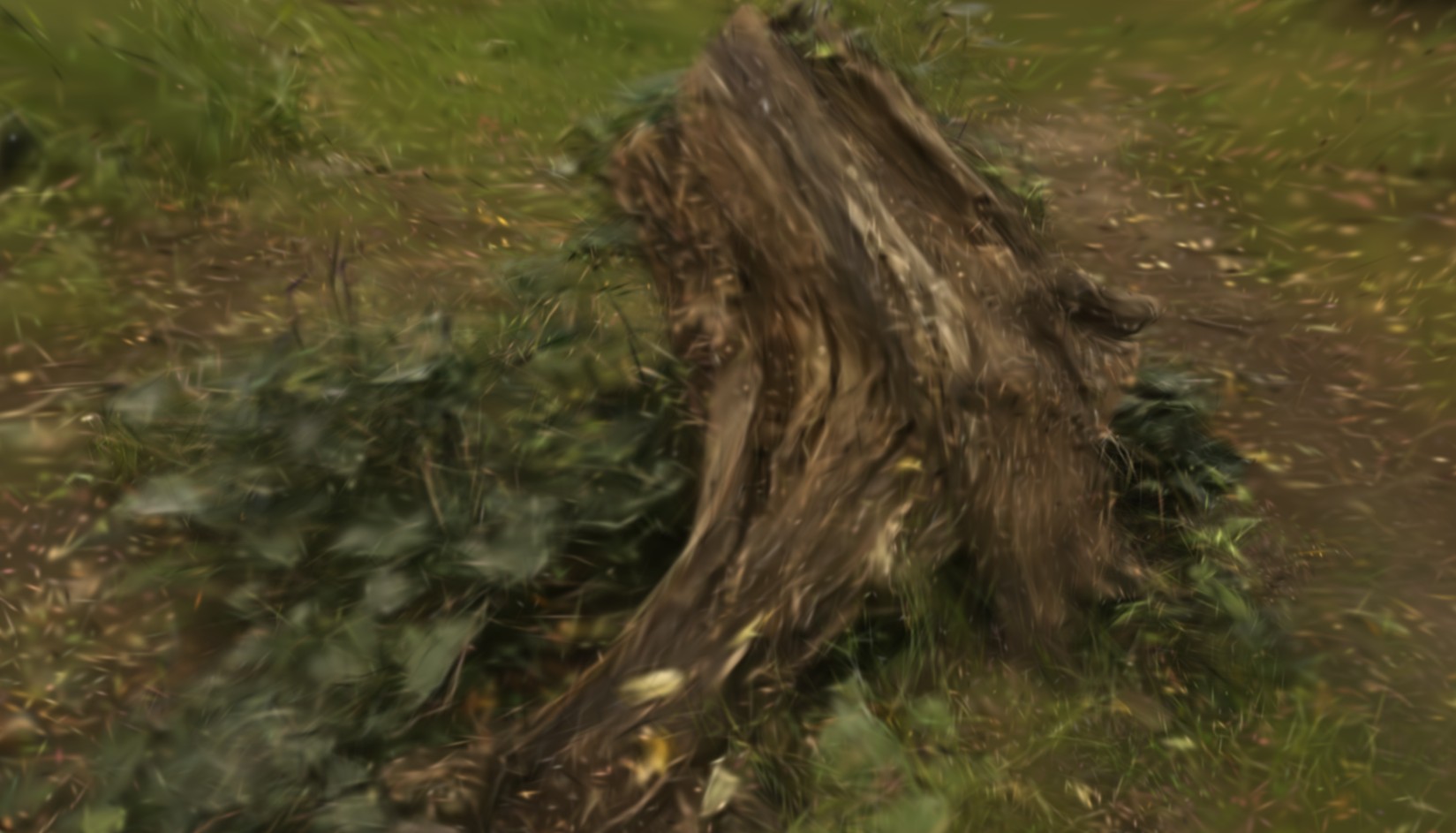}&
    \includegraphics[width=\fivewidth]{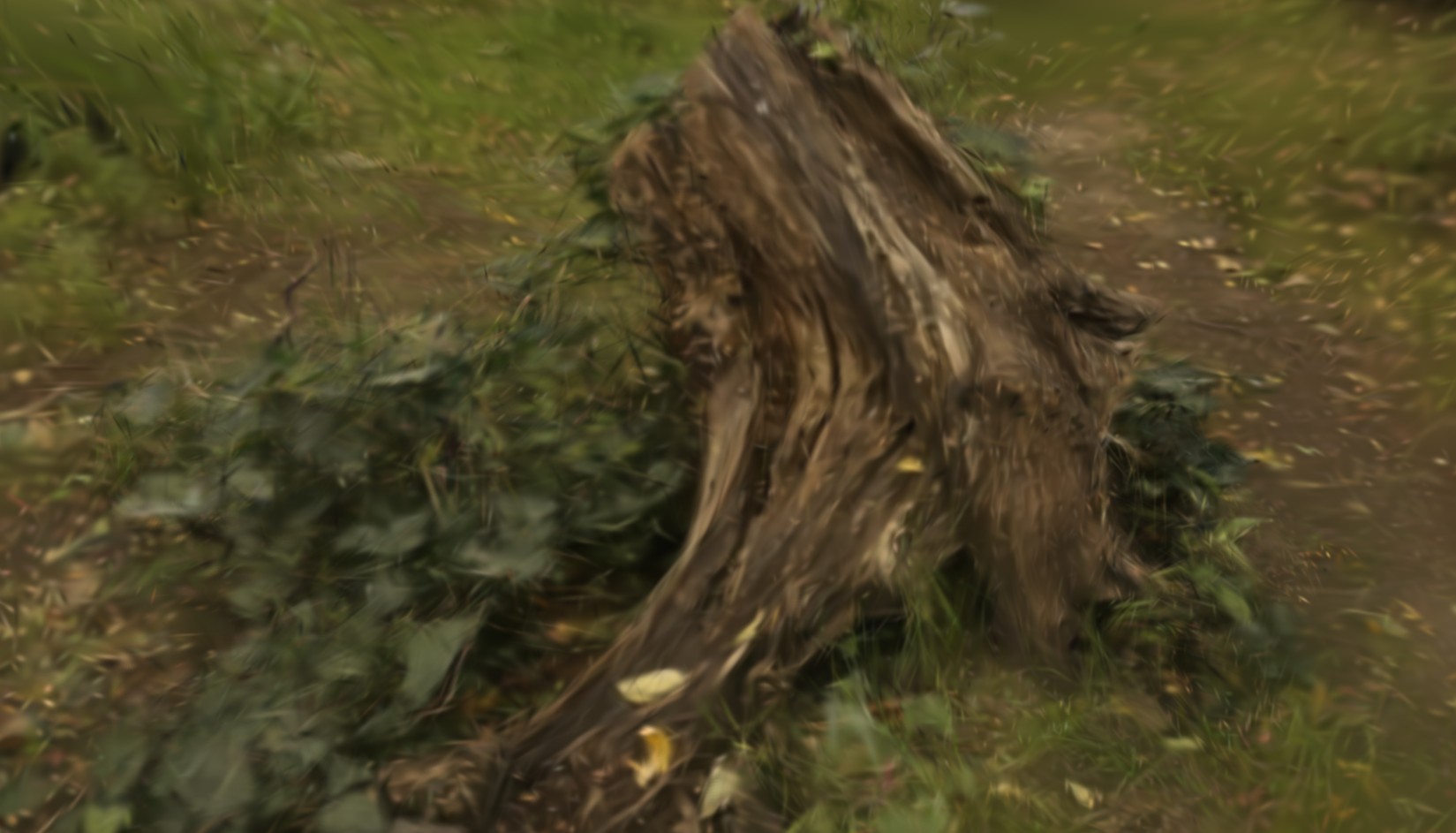}
    \\
    \includegraphics[width=\fivewidth]{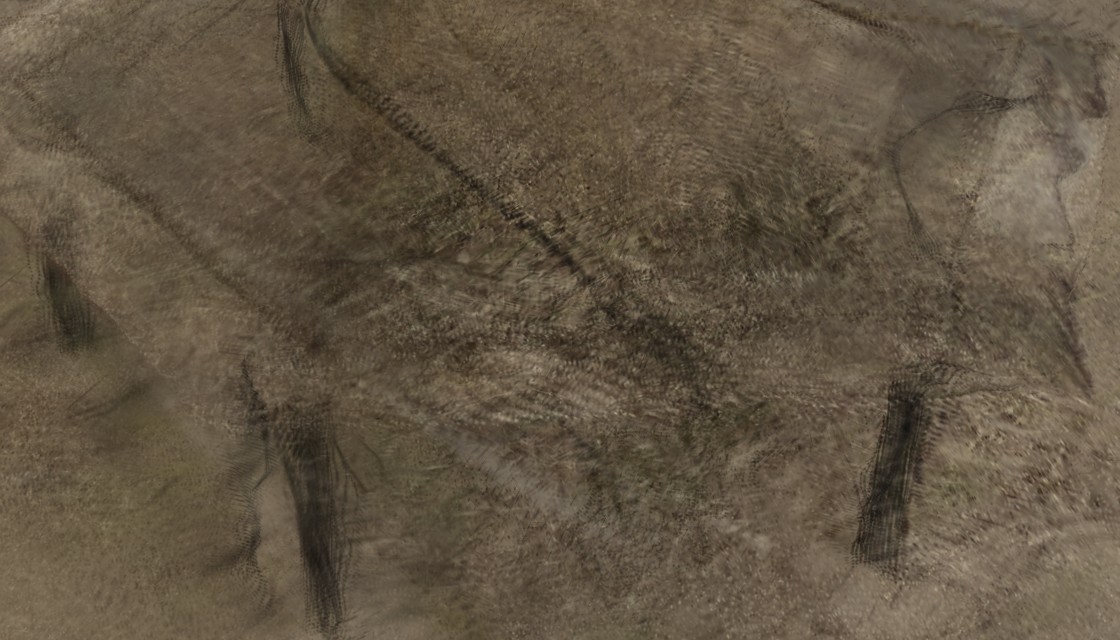}&
    \includegraphics[width=\fivewidth]{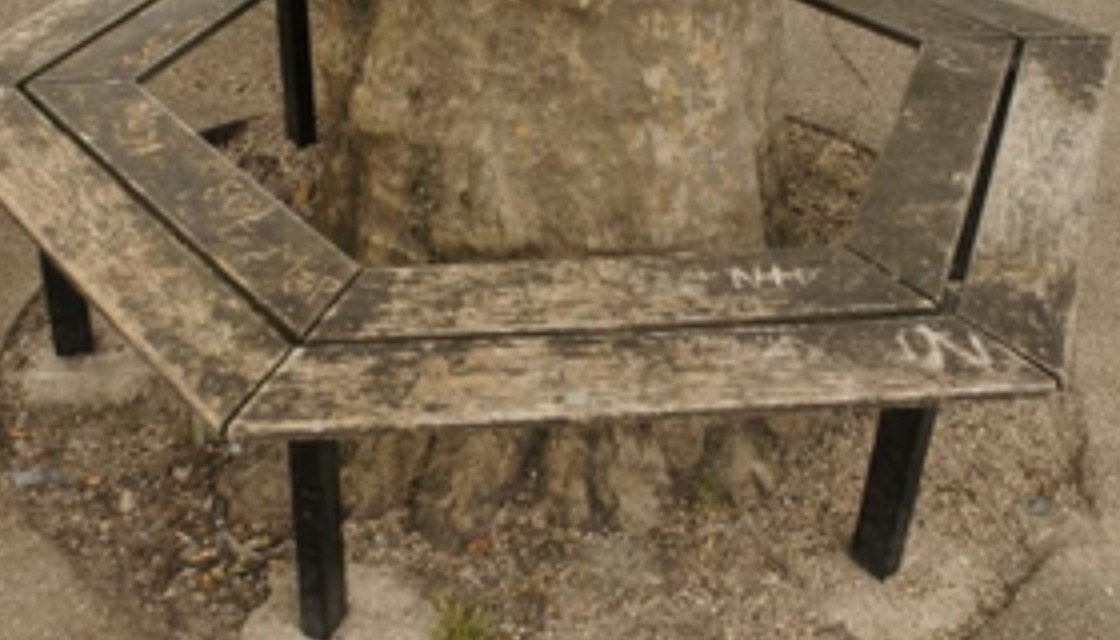}&
    \includegraphics[width=\fivewidth]{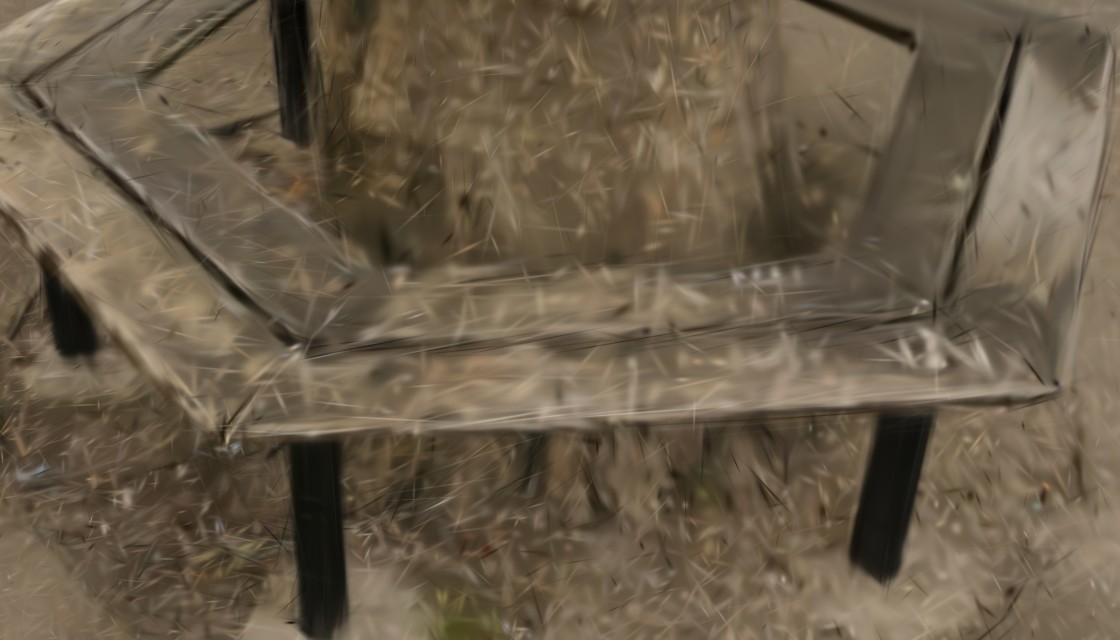}&
    \includegraphics[width=\fivewidth]{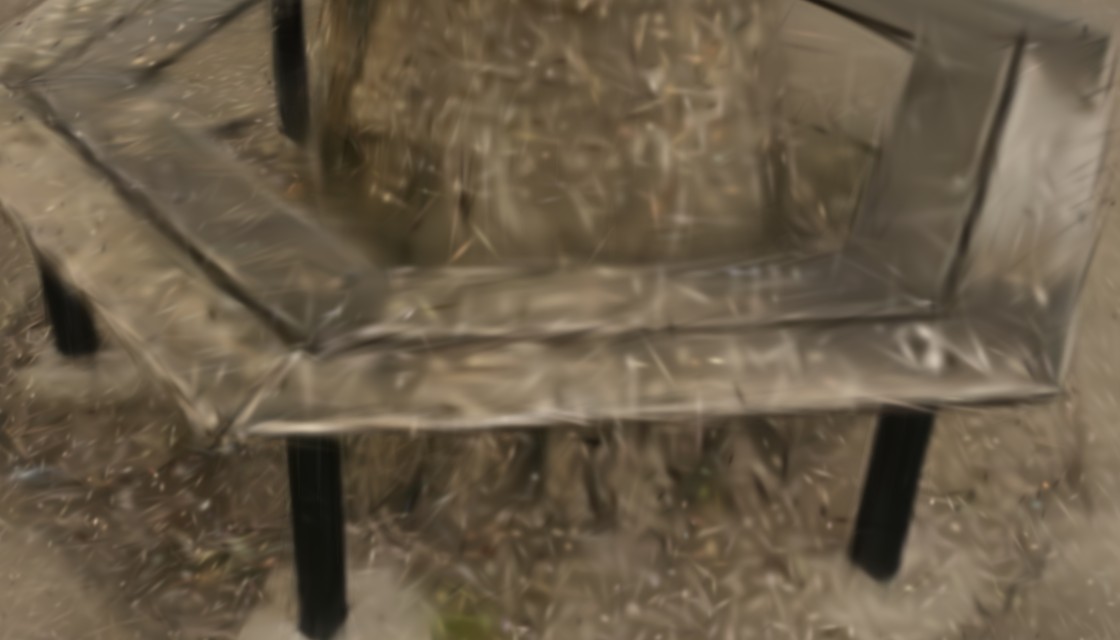}&
    \includegraphics[width=\fivewidth]{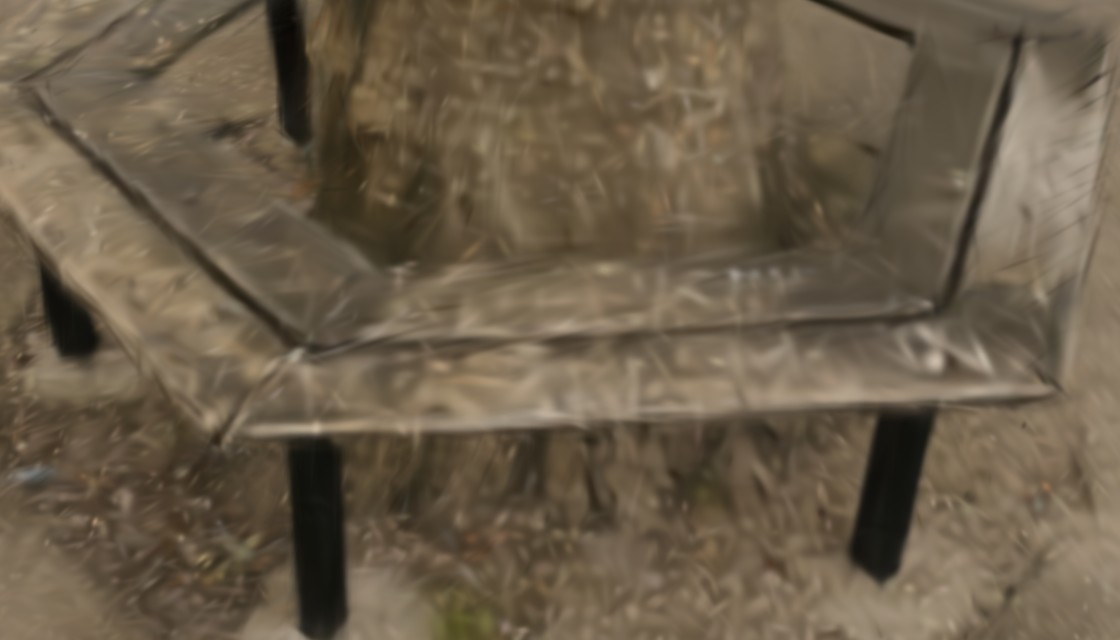}
    \\
    \begin{tabular}{@{}c@{}}\footnotesize SurfSplat\end{tabular} & 
    \begin{tabular}{@{}c@{}}\footnotesize GT \\ $\times 1$ resolution\end{tabular} & 
    \begin{tabular}{@{}c@{}}\footnotesize RS+\\ \footnotesize gsplat\end{tabular} & 
    \begin{tabular}{@{}c@{}}\footnotesize RS+GGGS\\ \footnotesize +EANC\end{tabular} & 
    \begin{tabular}{@{}c@{}}\footnotesize RS+GOF\\ \footnotesize +EANC\end{tabular} 
    \end{tabular}
    
    \caption{\textbf{High-Resolution Comparison on DL3DV test set and Mip-NeRF 360.} Color renderings at $\times 4$ resolution. SurfSplat fails to align its predicted Gaussians. RS+gsplat is trained without the 3D smoothing filter thus exhibiting artifacts in high-frequency regions such as the bench's bars, and details on the stump and tree. Our method limits the predicted Gaussians' extent avoiding spurious elongated primitives.}
    \label{fig:hrrc_4x_comparison}
\end{figure*}

\newcommand{\imgw}{0.2125\textwidth} 

\newcommand{\vcenterimg}[1]{\begin{tabular}{@{}c@{}}\includegraphics[width=\imgw]{#1}\end{tabular}}

\newcommand{\vcenterrot}[1]{\begin{tabular}{@{}c@{}}\rotatebox[origin=c]{90}{\scriptsize \begin{tabular}{@{}c@{}}#1\end{tabular}}\end{tabular}}

\begin{figure*}[t]
    \centering
    \setlength{\tabcolsep}{0.1em}

    \begin{tabular}{c c c c c}
    & \scriptsize Barn & \scriptsize Caterpillar & \scriptsize Ignatius & \scriptsize Truck \\[0.1em]
    
    \vcenterrot{Color GT} & 
    \vcenterimg{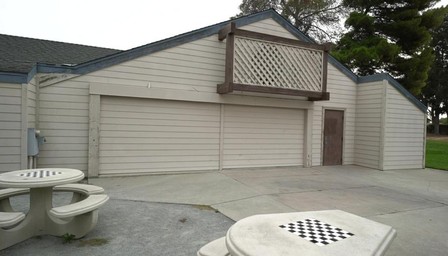}&
    \vcenterimg{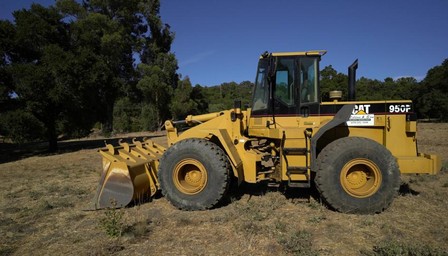}&
    \vcenterimg{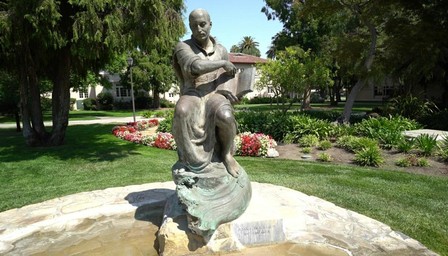}&
    \vcenterimg{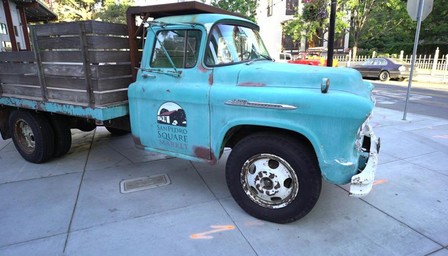} \\[0.2em]
    
    \vcenterrot{SurfSplat} & 
    \vcenterimg{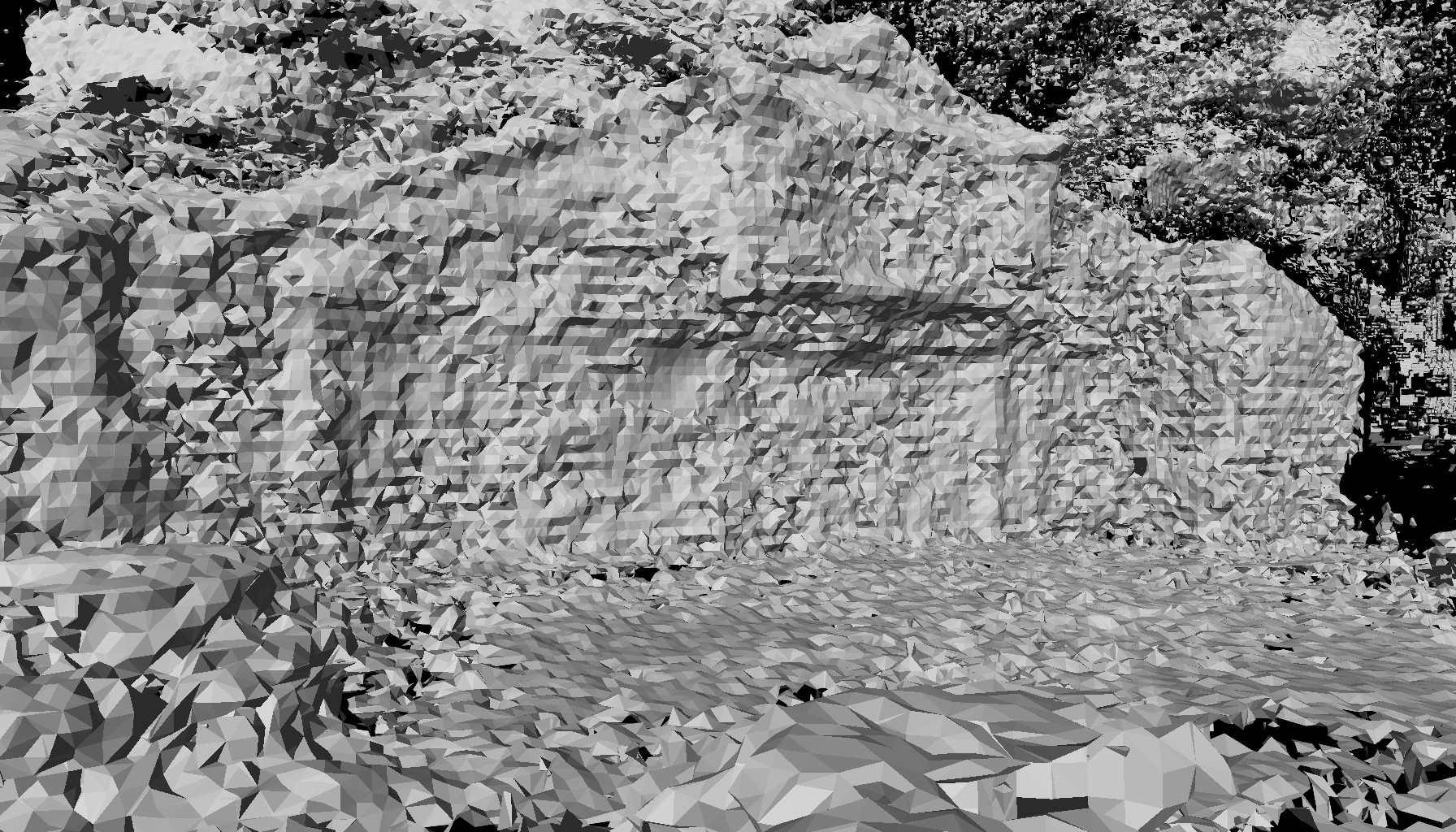}&
    \vcenterimg{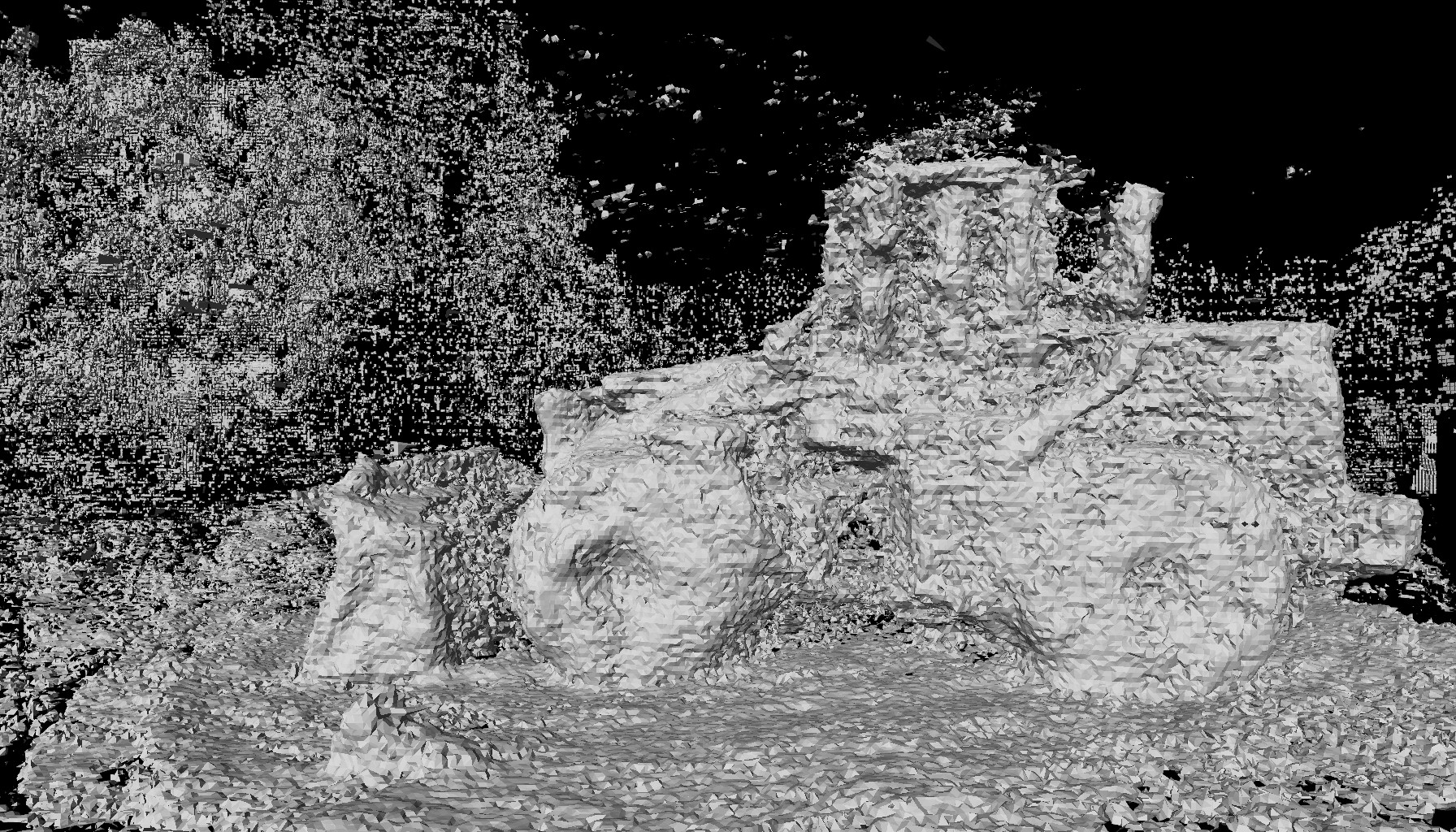}&
    \vcenterimg{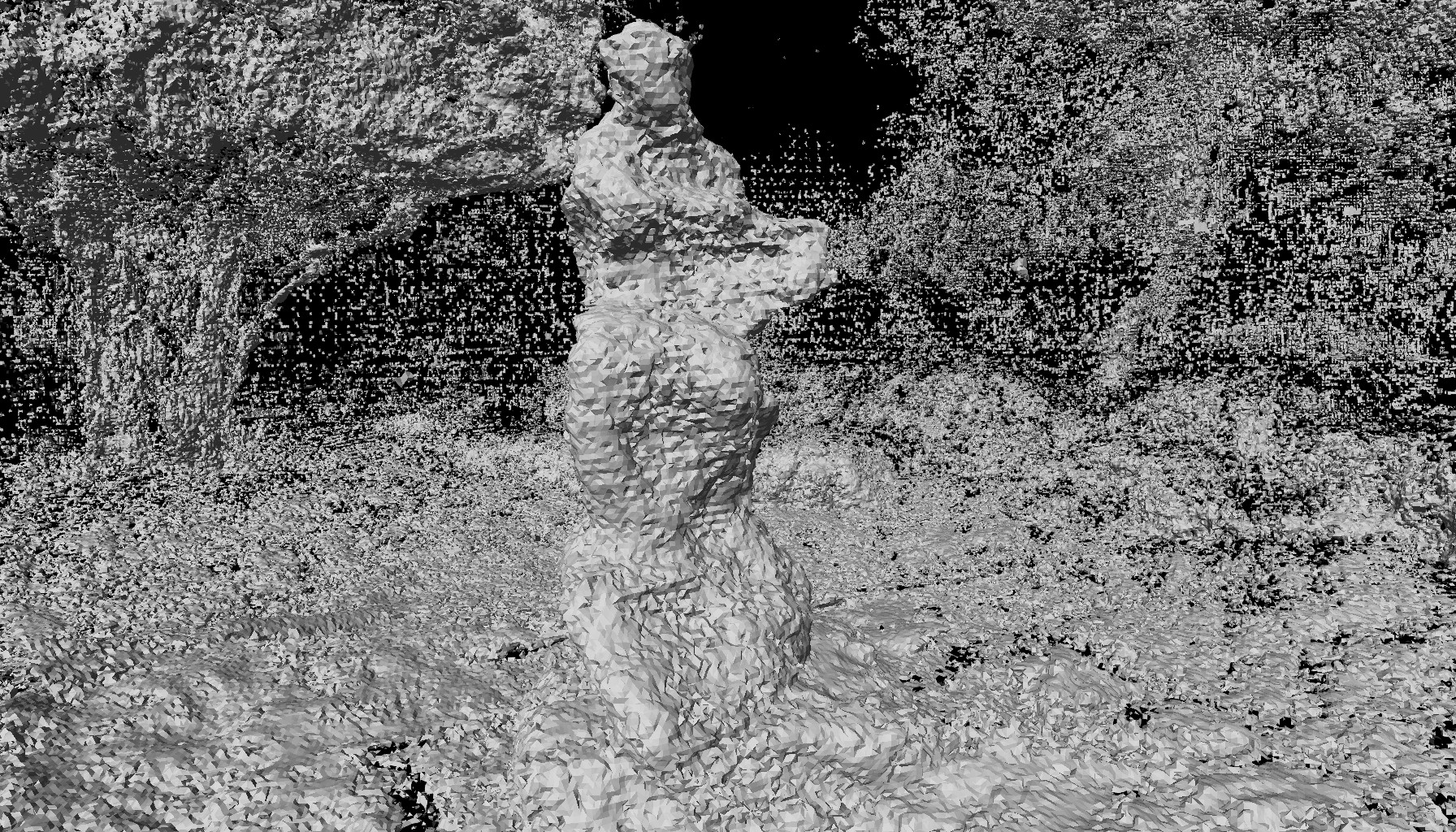}&
    \vcenterimg{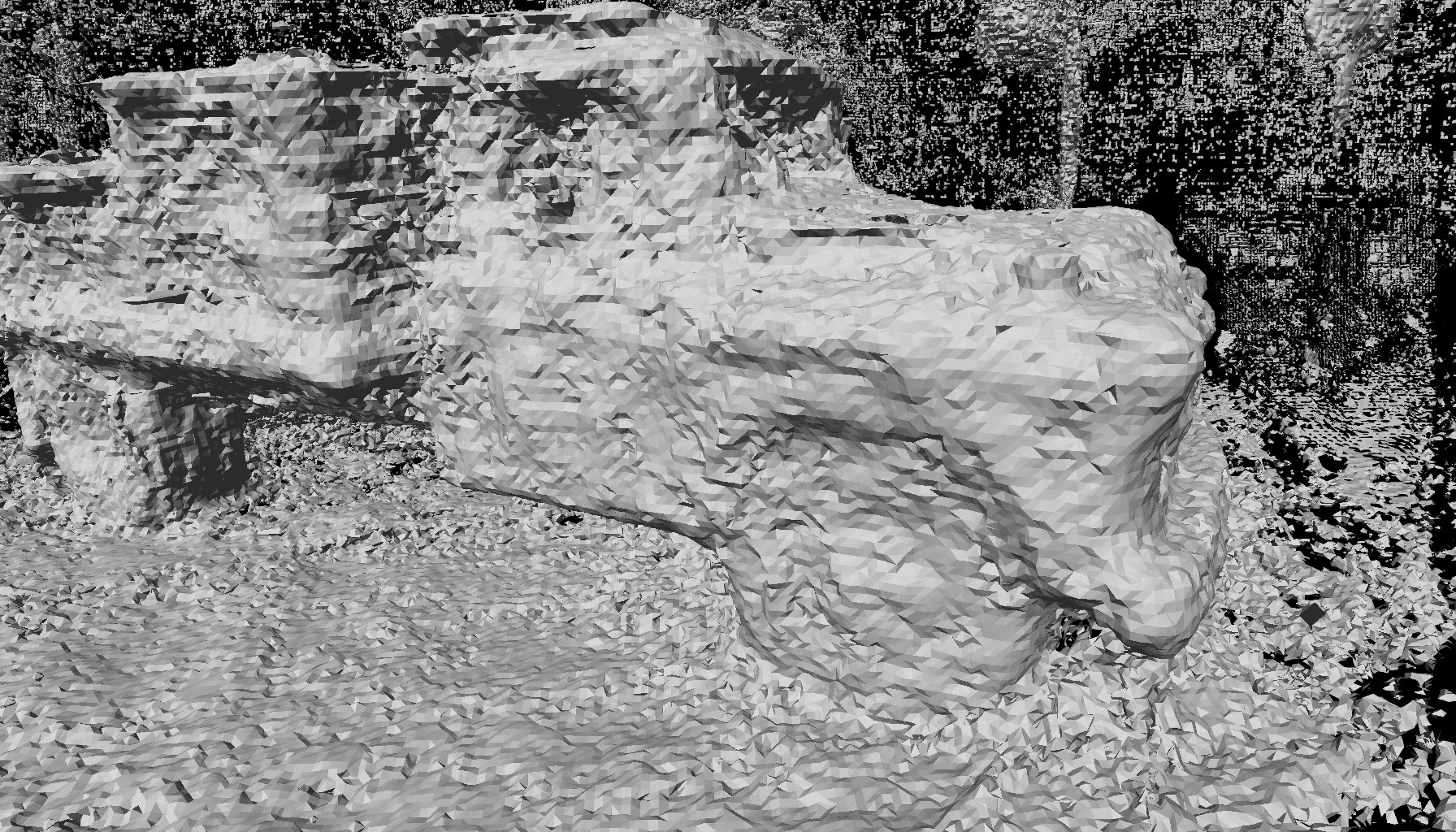} \\[0.2em]
    
    \vcenterrot{DA3} & 
    \vcenterimg{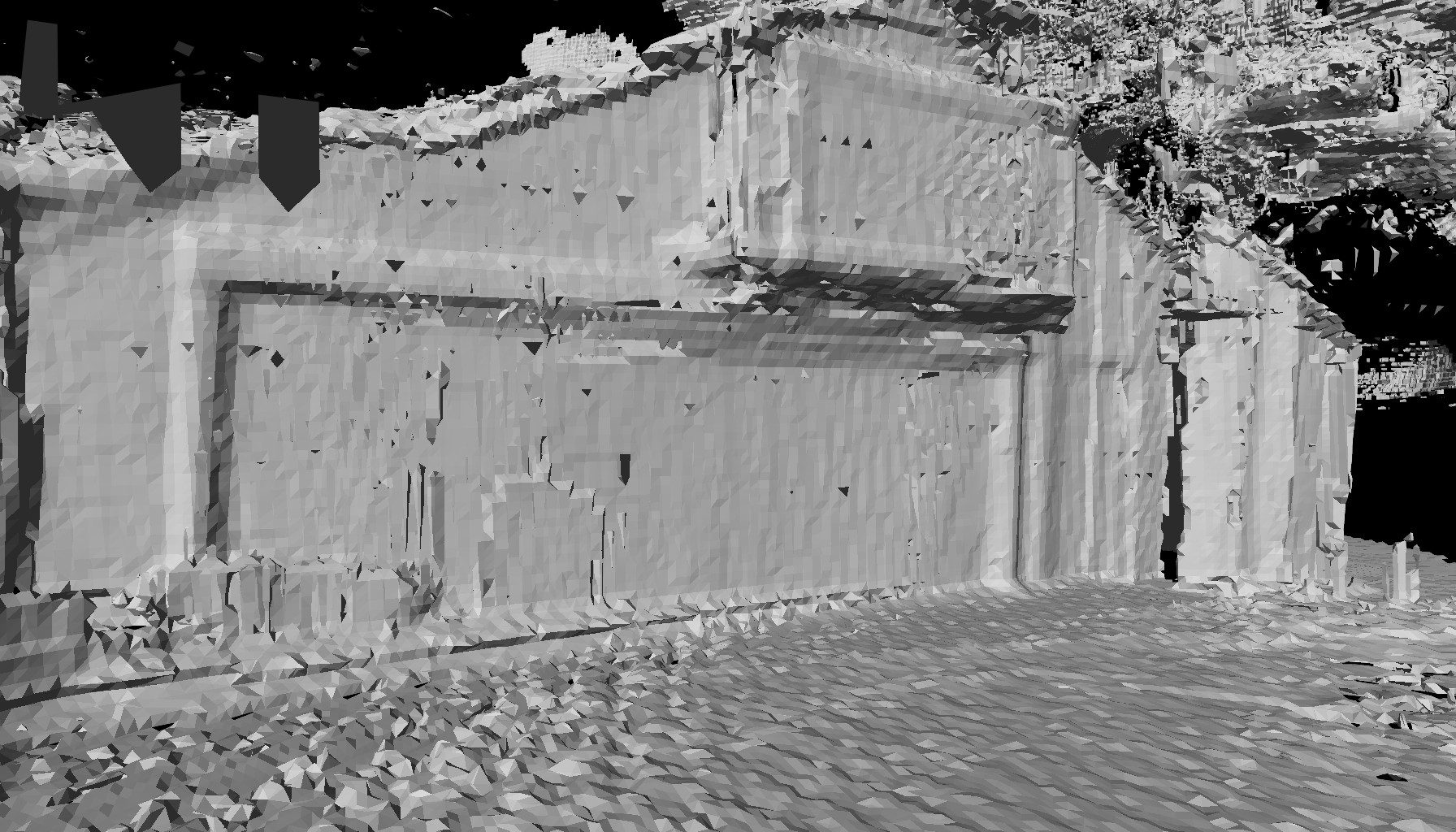}&
    \vcenterimg{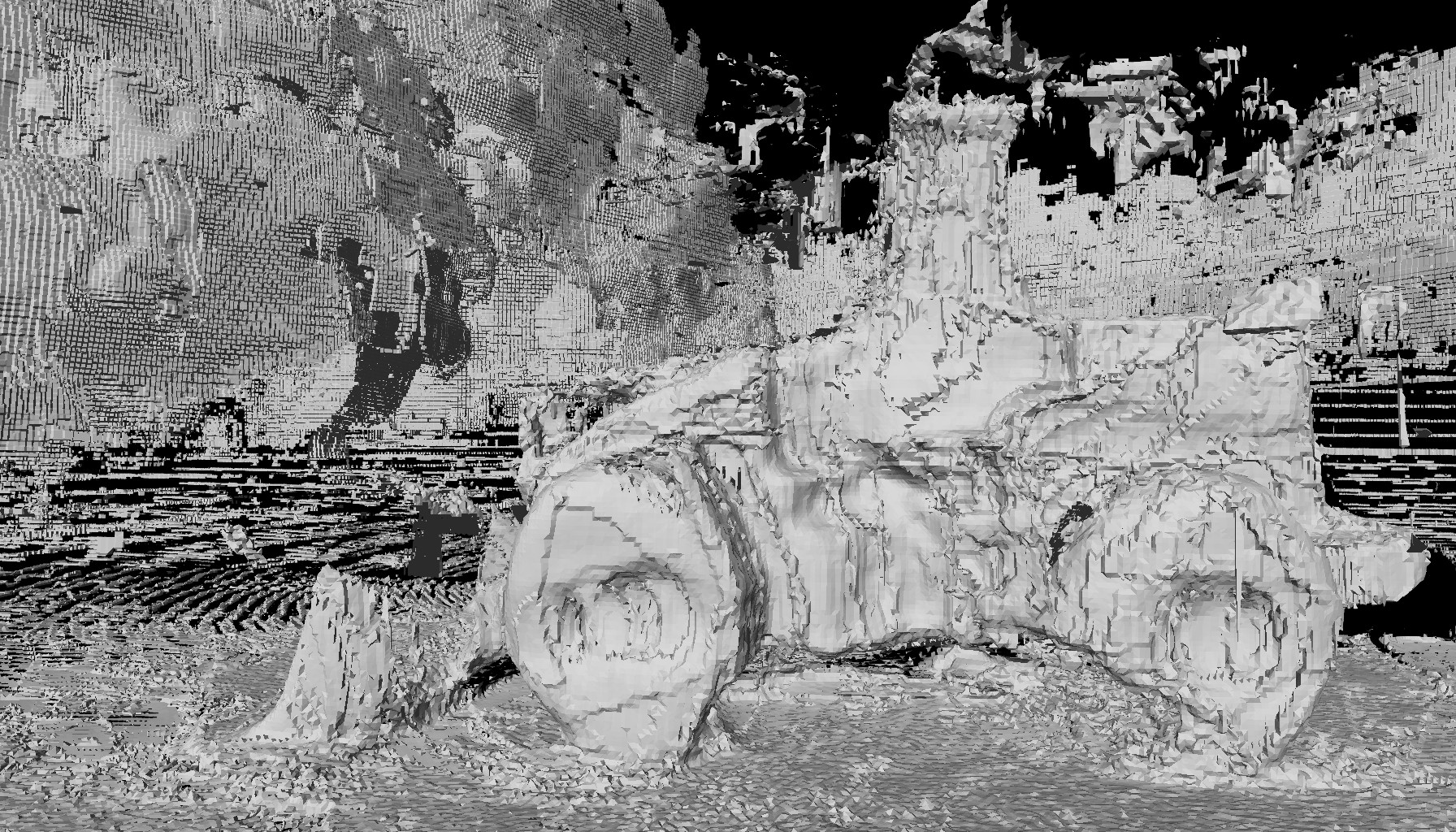}&
    \vcenterimg{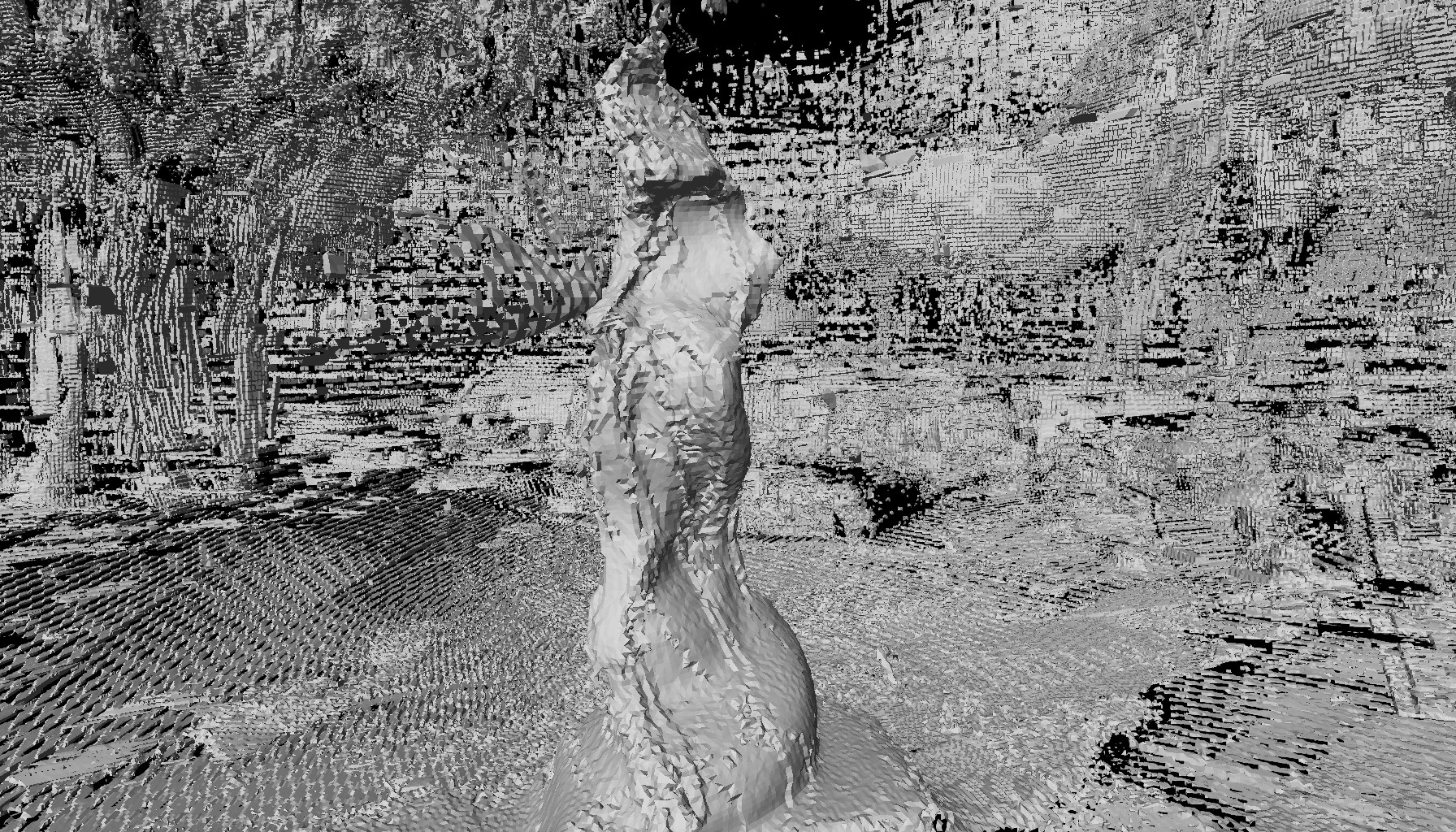}&
    \vcenterimg{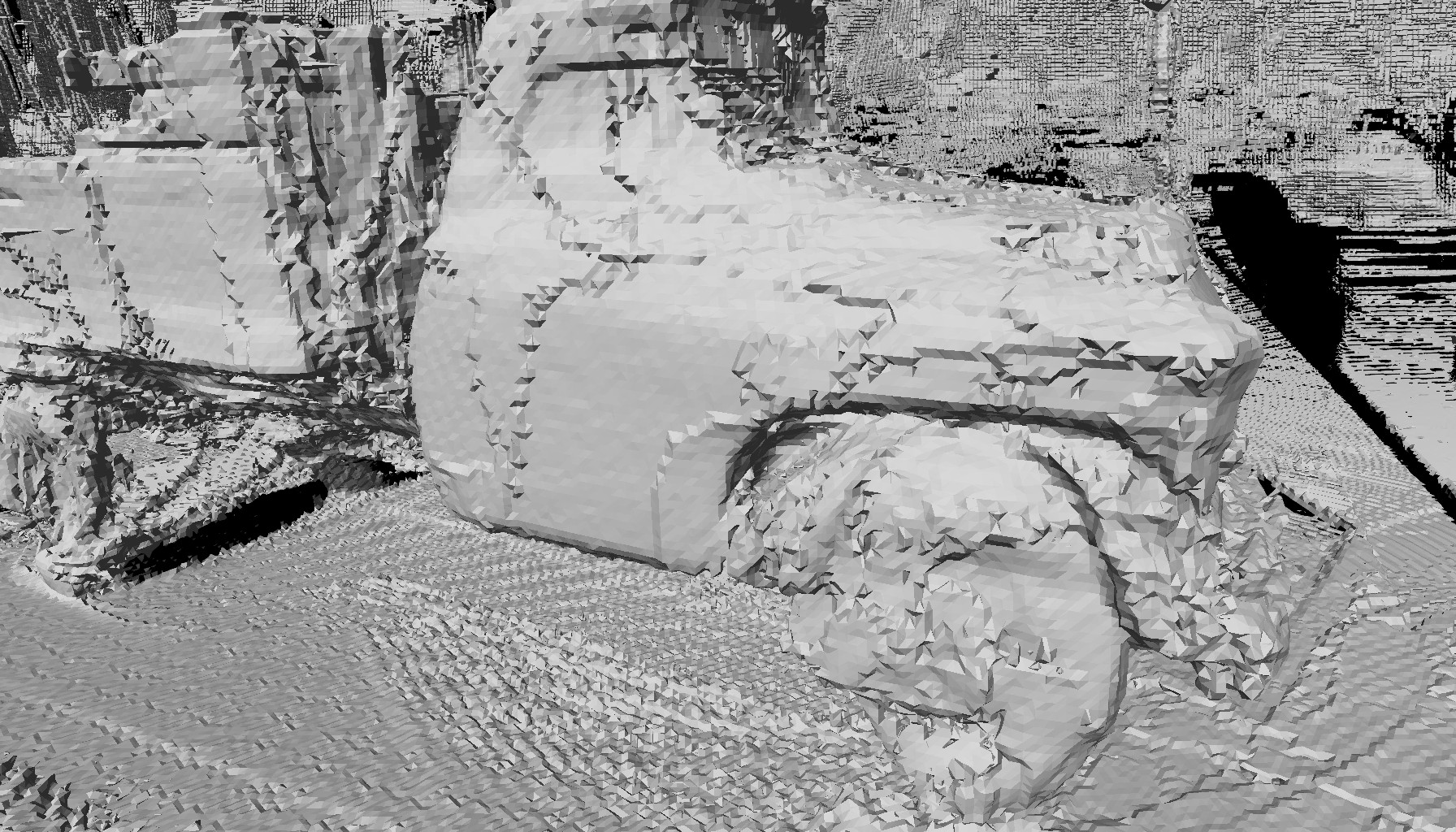} \\[0.2em]
    
    \vcenterrot{RS+\\gsplat} & 
    \vcenterimg{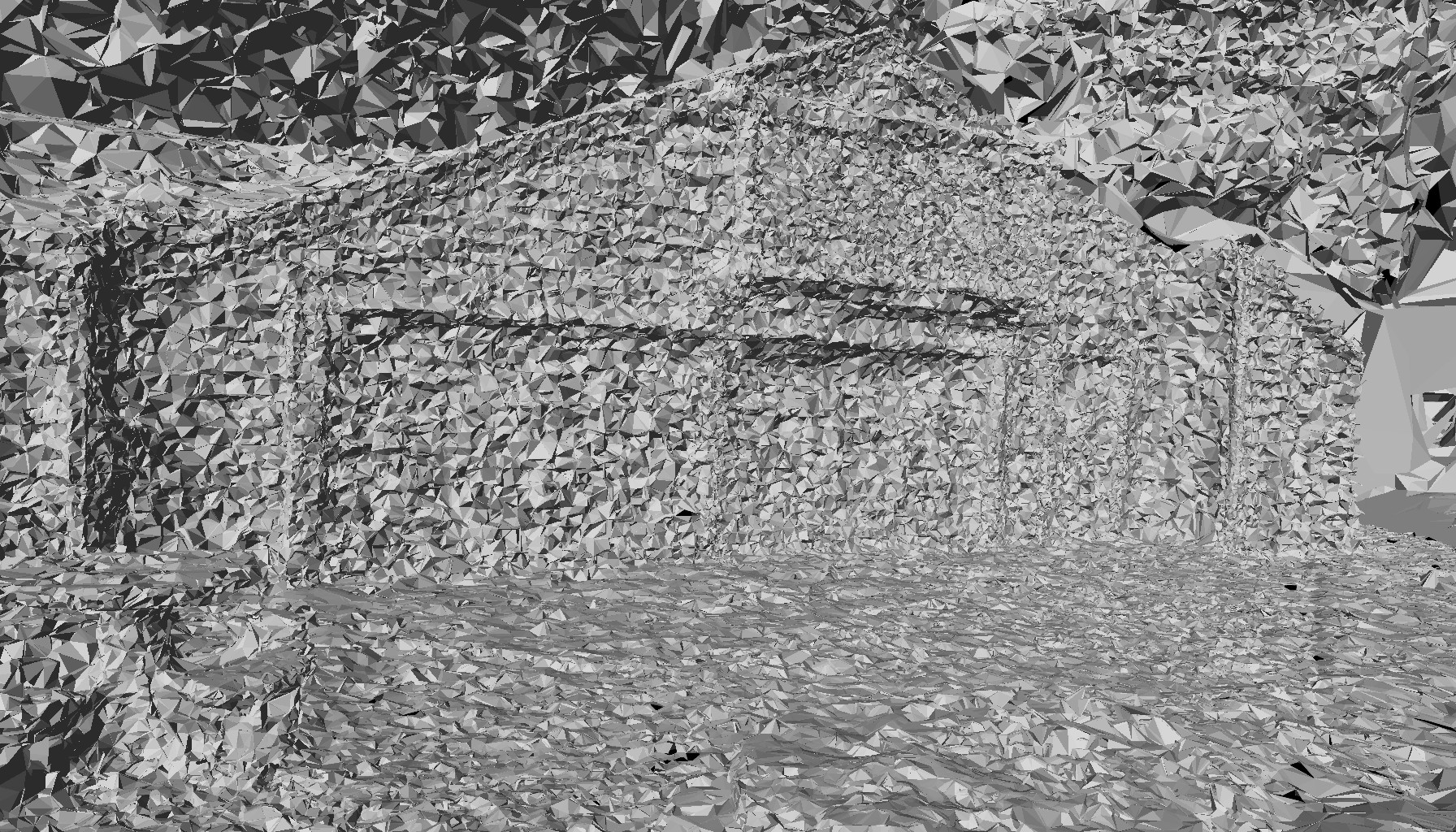}&
    \vcenterimg{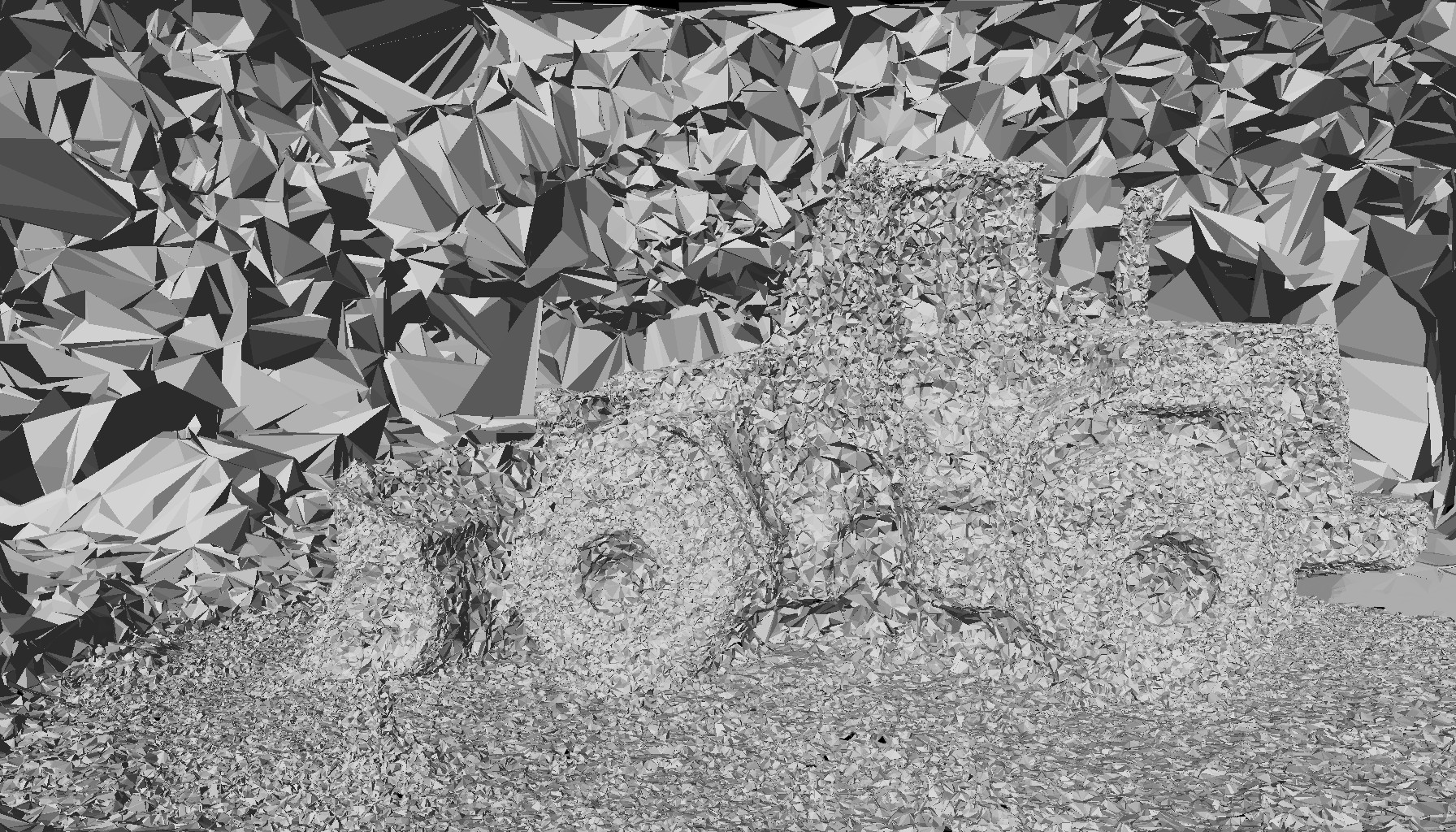}&
    \vcenterimg{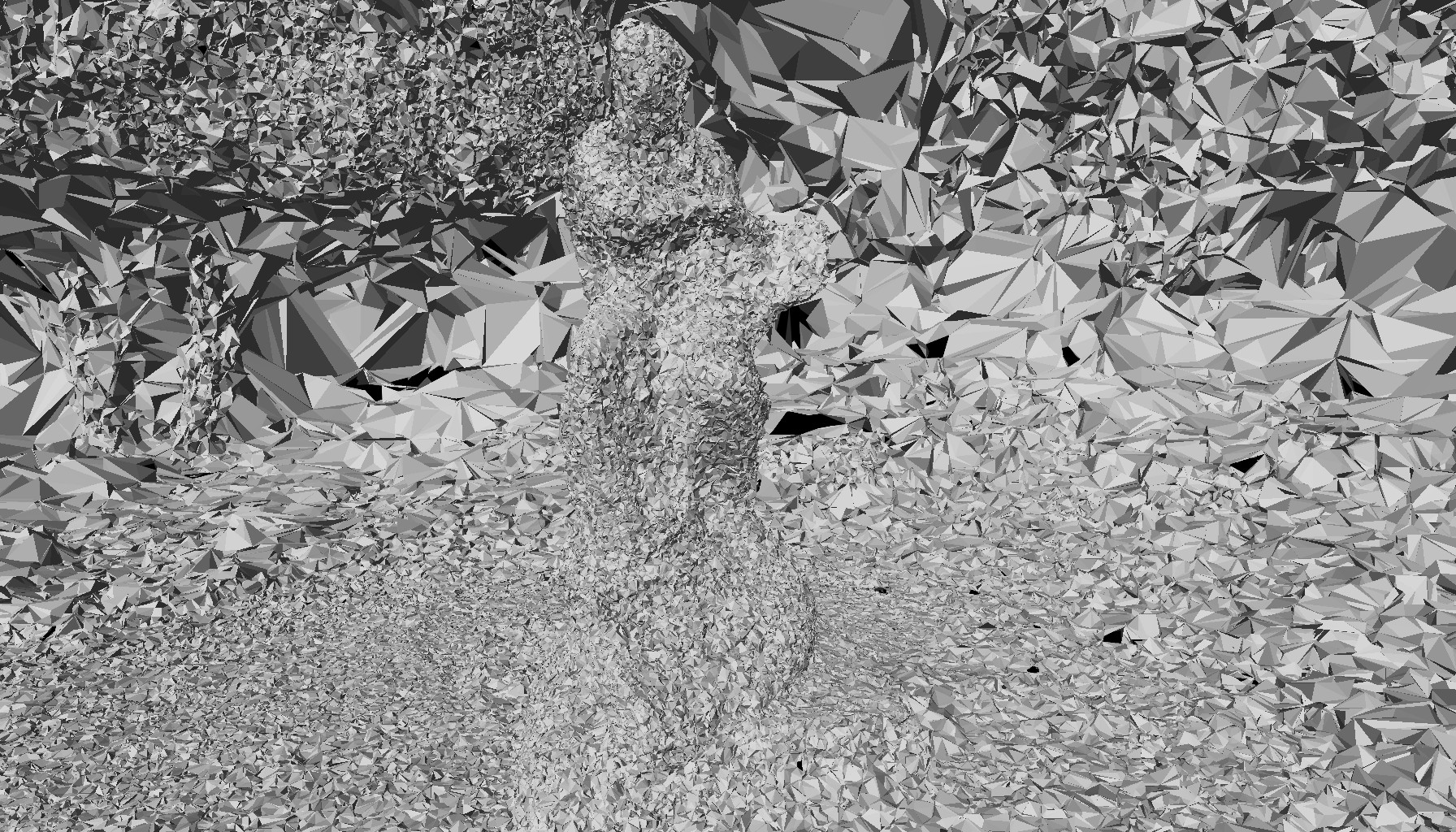}&
    \vcenterimg{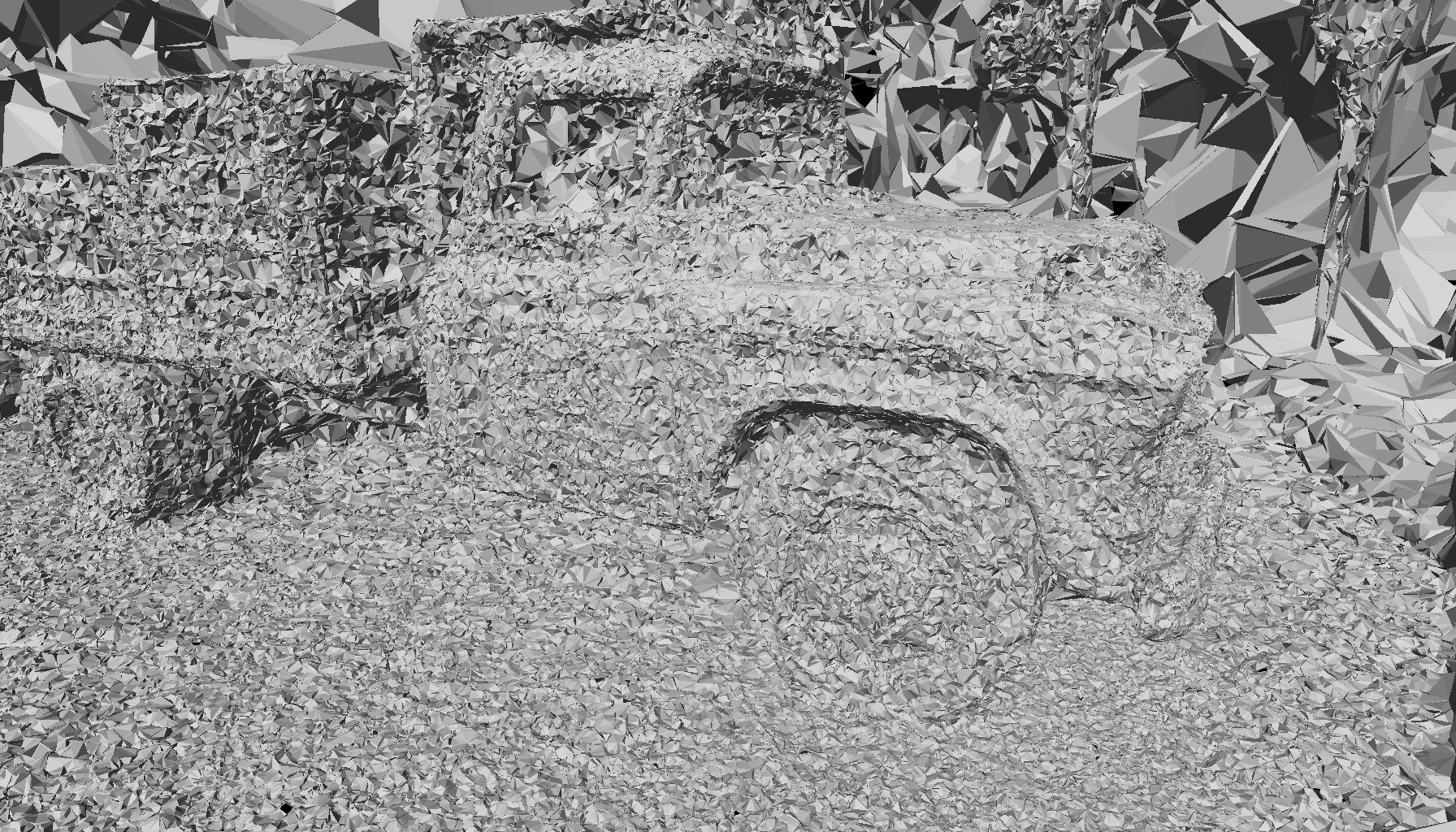} \\[0.2em]
    
    \vcenterrot{RS+GGGS\\+EANC} & 
    \vcenterimg{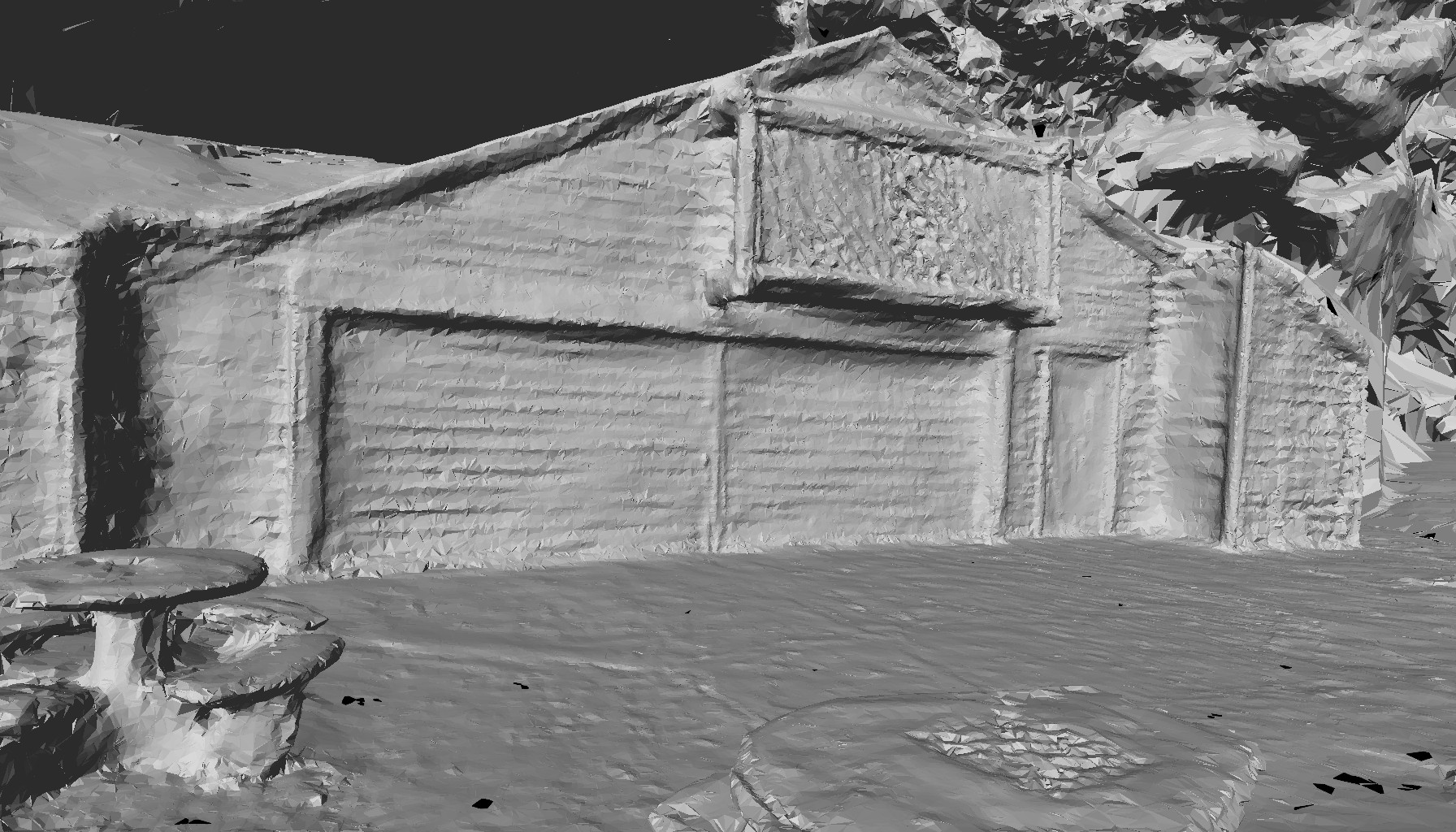}&
    \vcenterimg{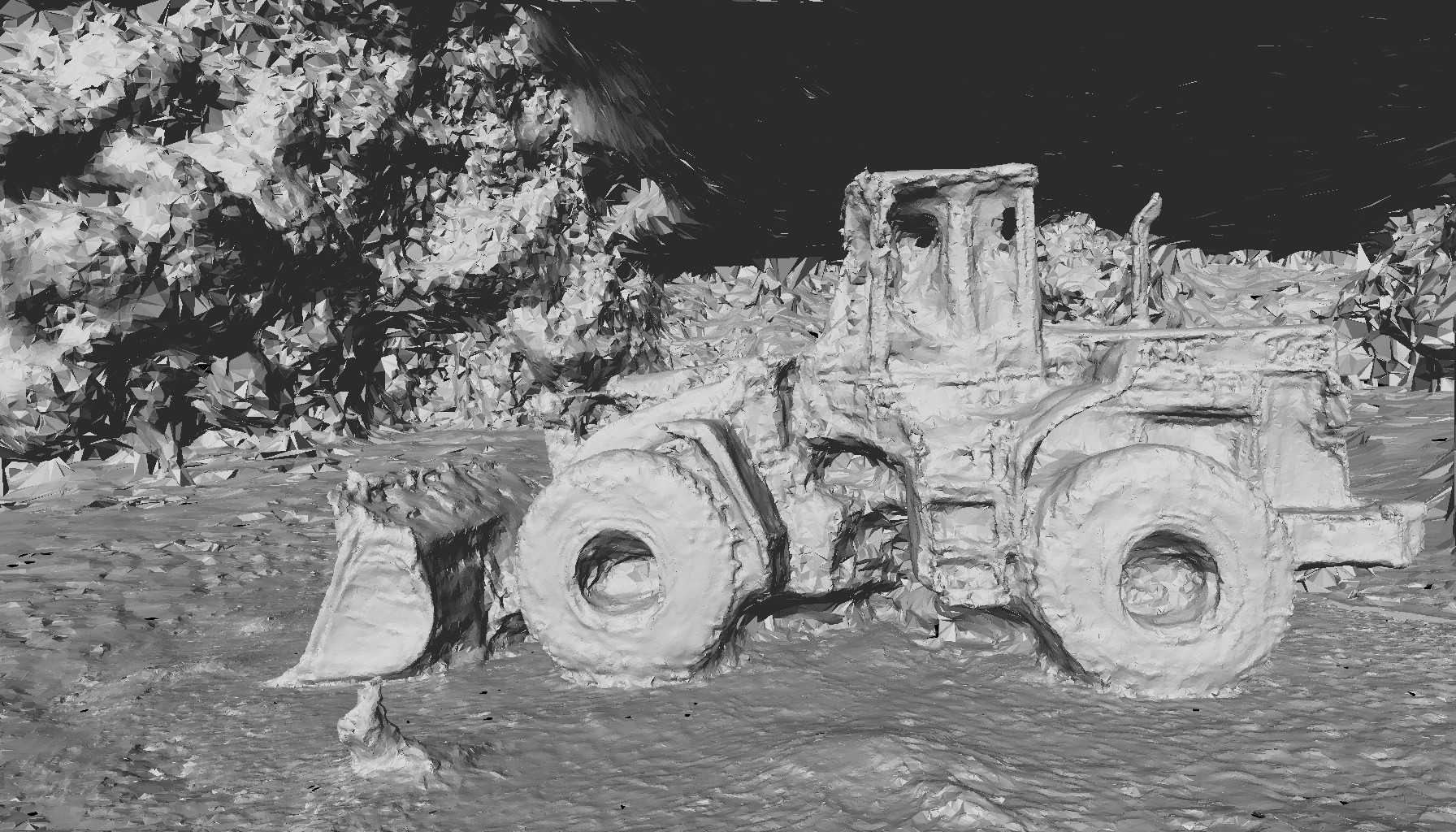}&
    \vcenterimg{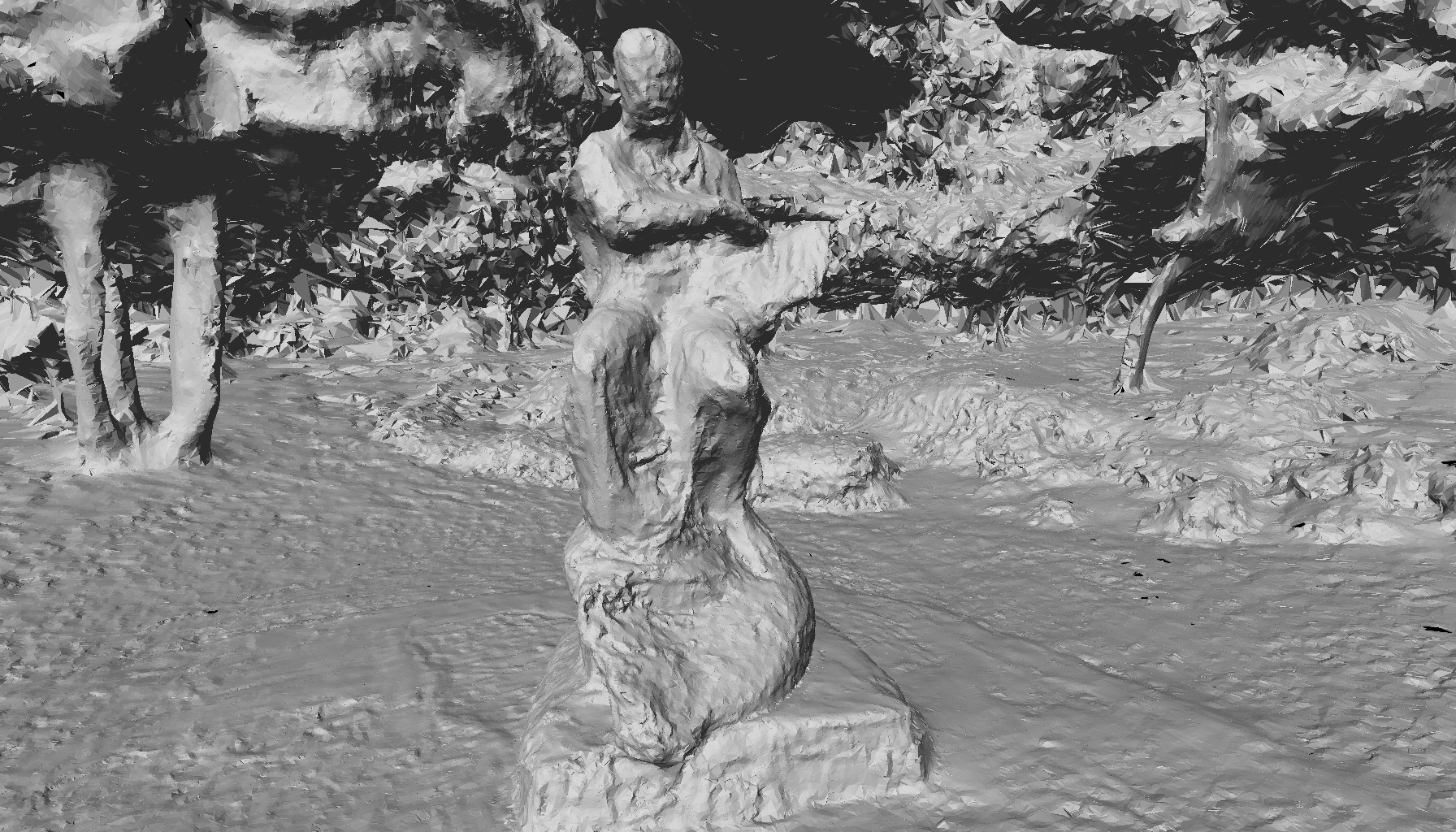}&
    \vcenterimg{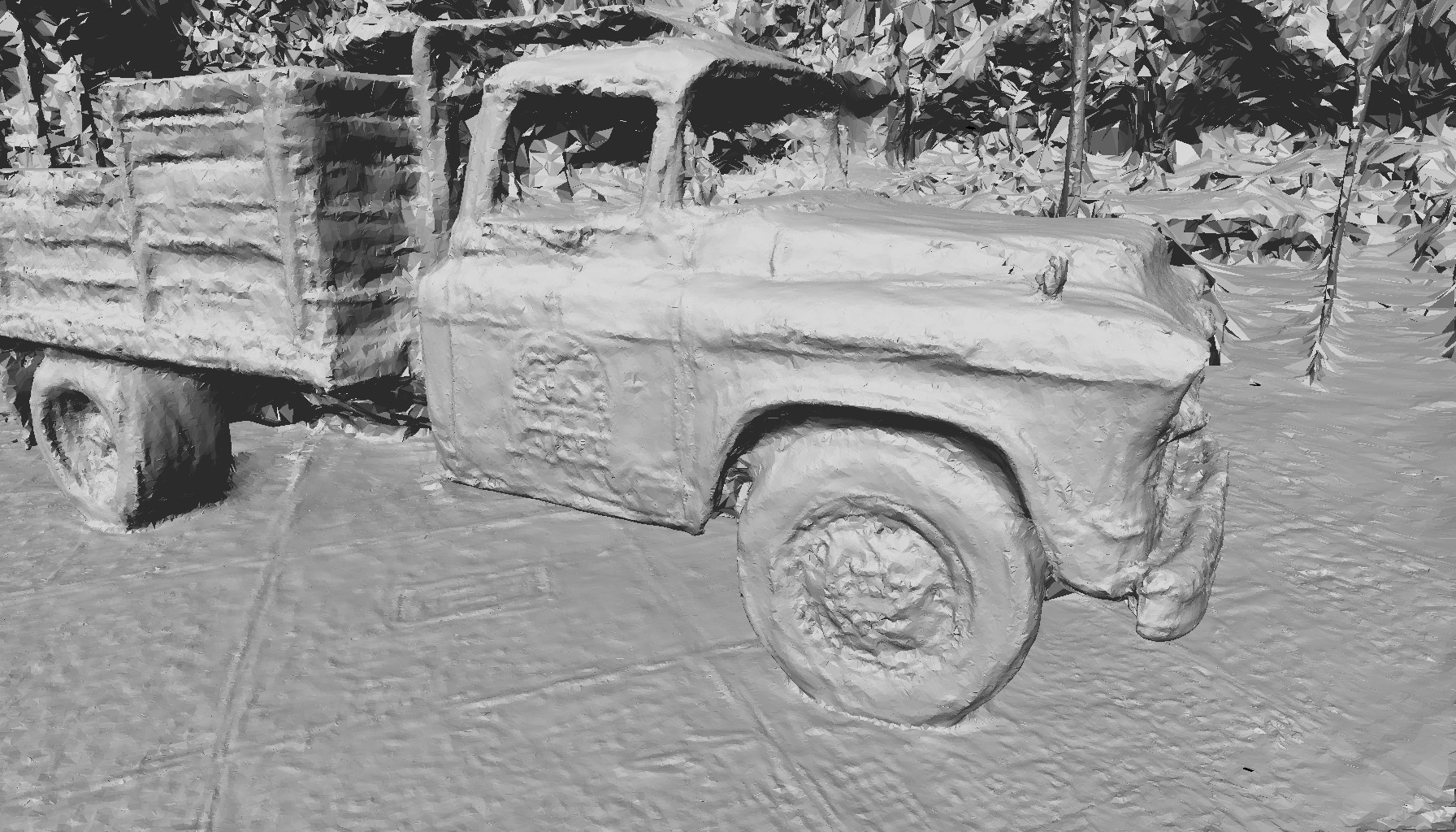} \\[0.2em]
    
    \vcenterrot{RS+GOF\\+EANC} & 
    \vcenterimg{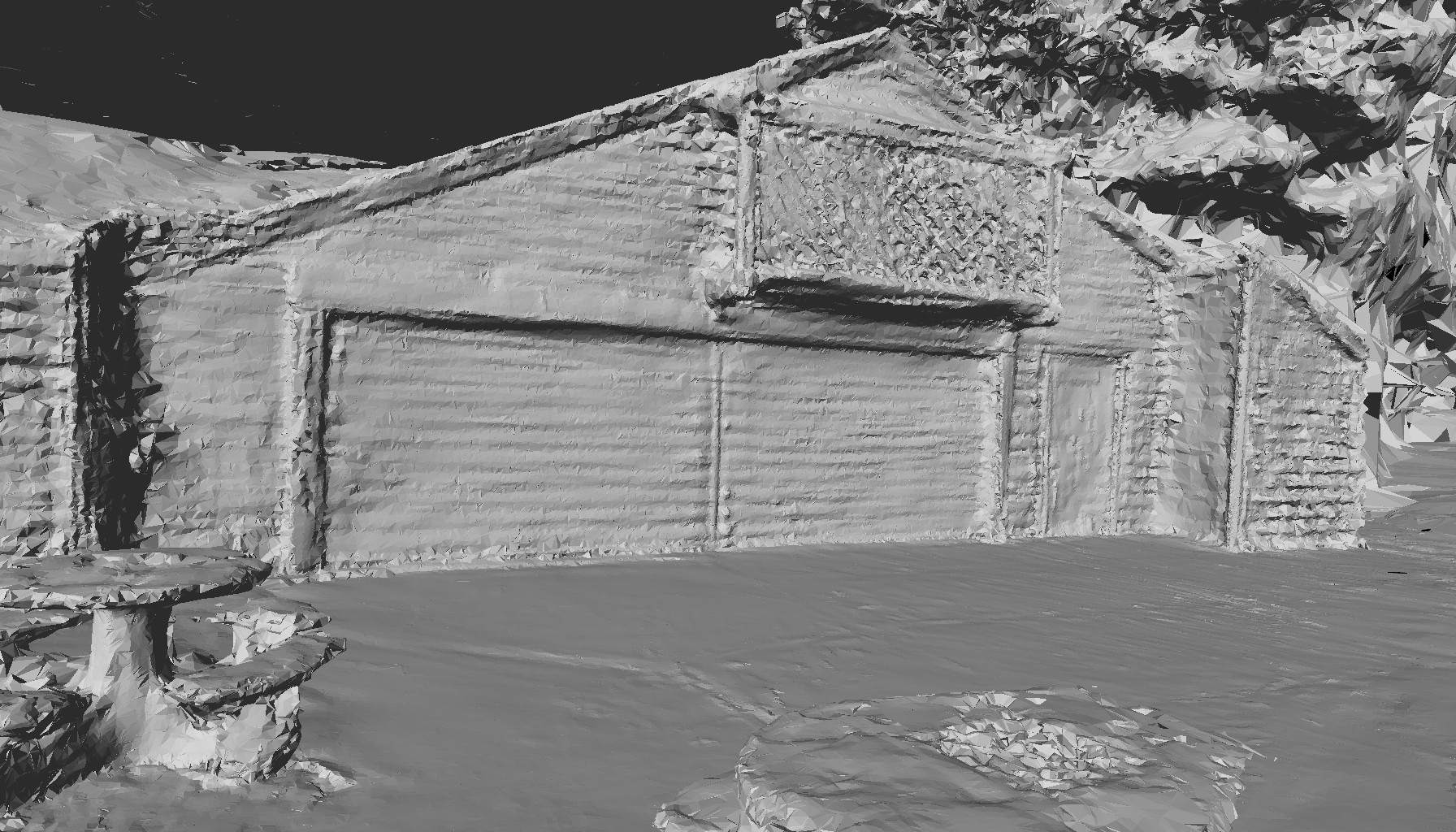}&
    \vcenterimg{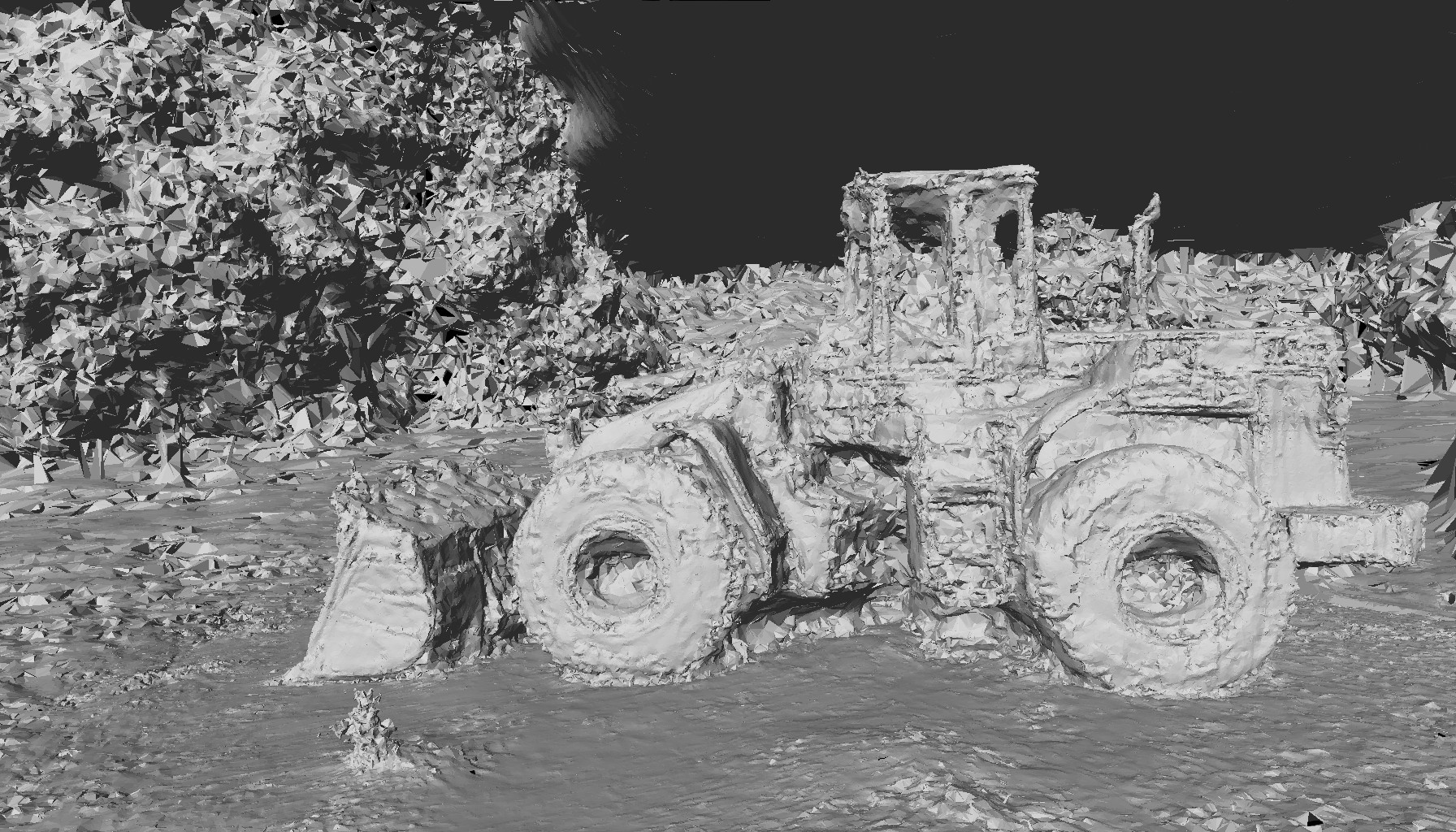}&
    \vcenterimg{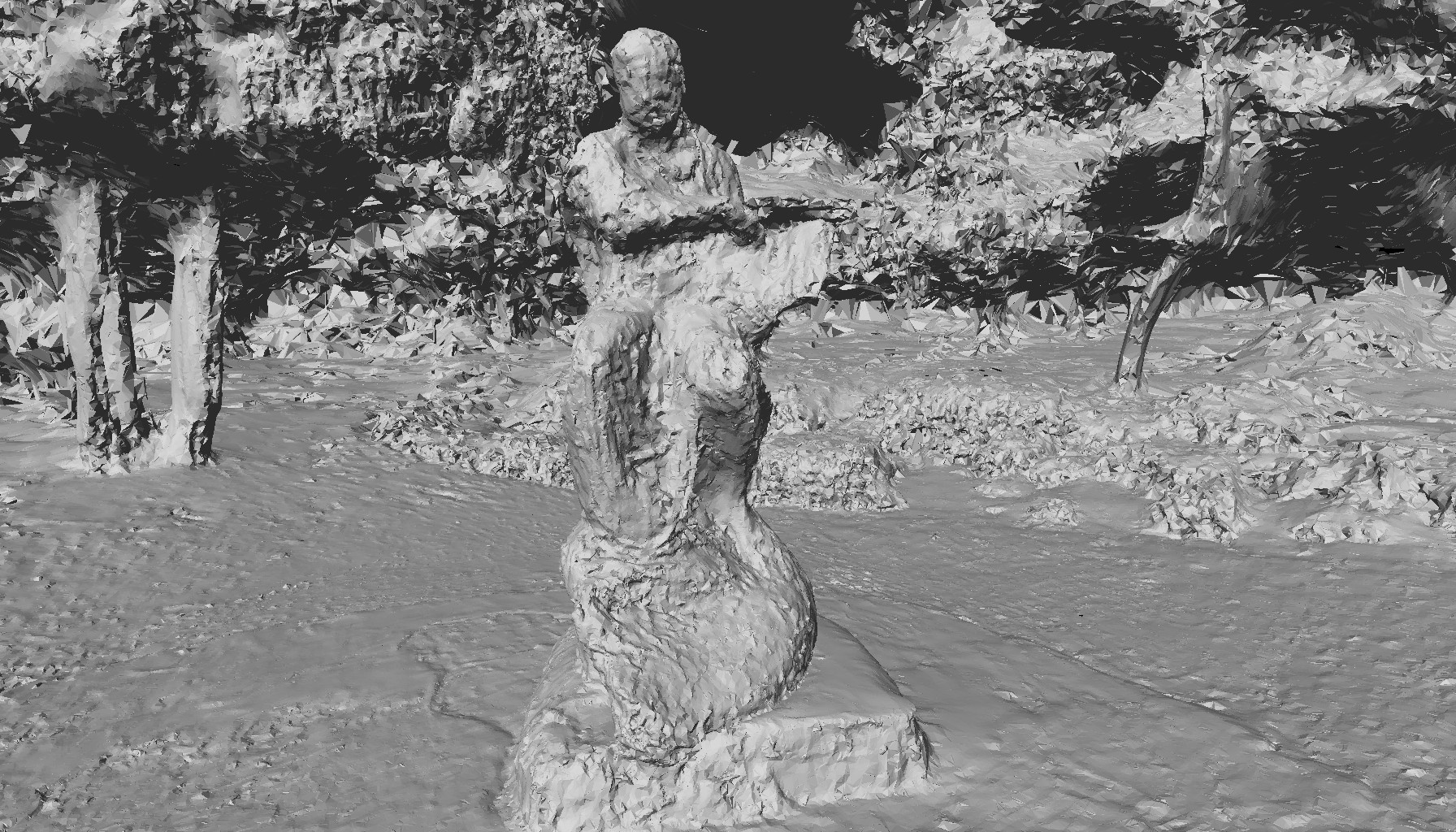}&
    \vcenterimg{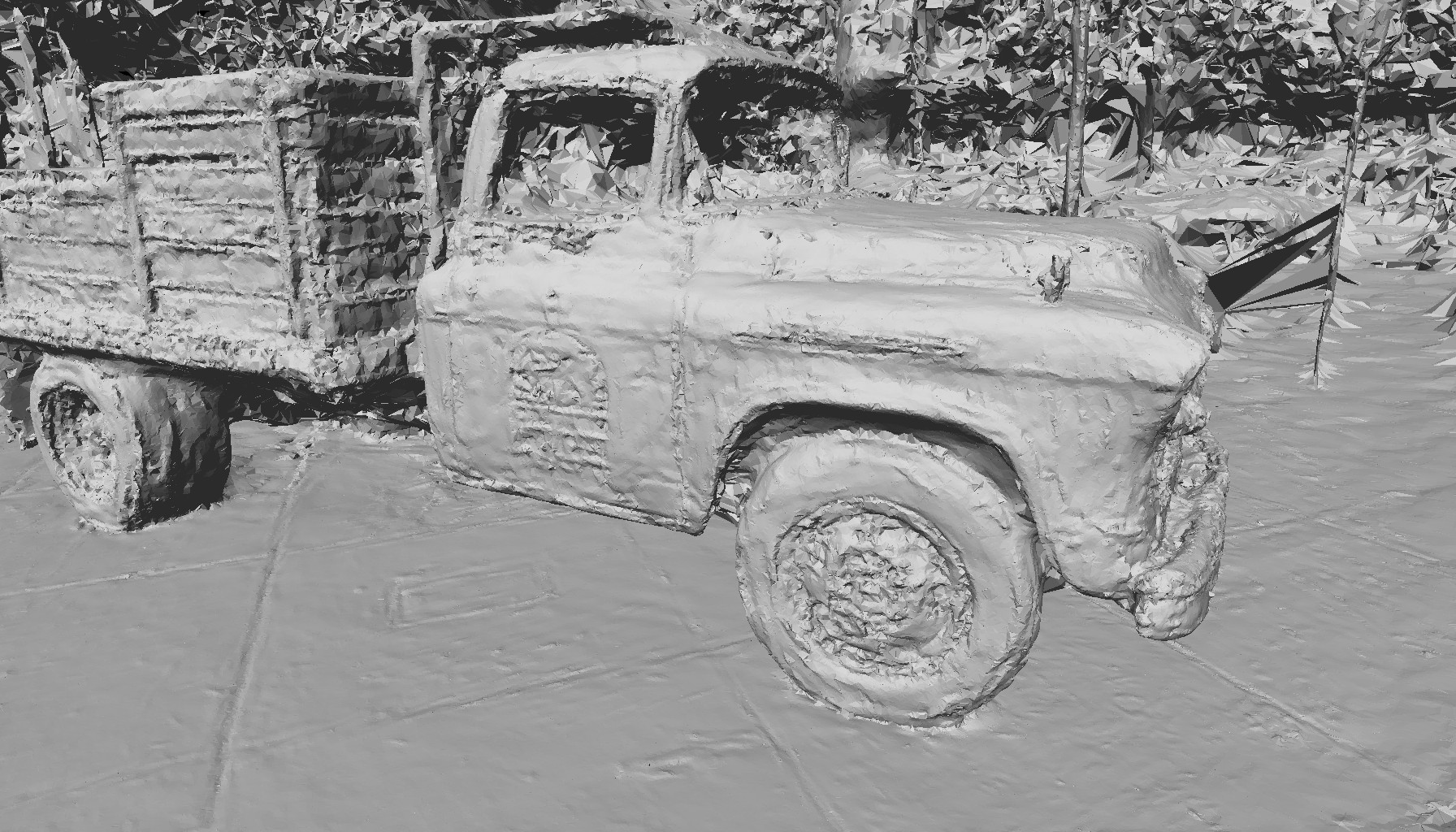} \\[0.2em]
    
    \vcenterrot{GGGS Opt.\\4k steps} & 
    \vcenterimg{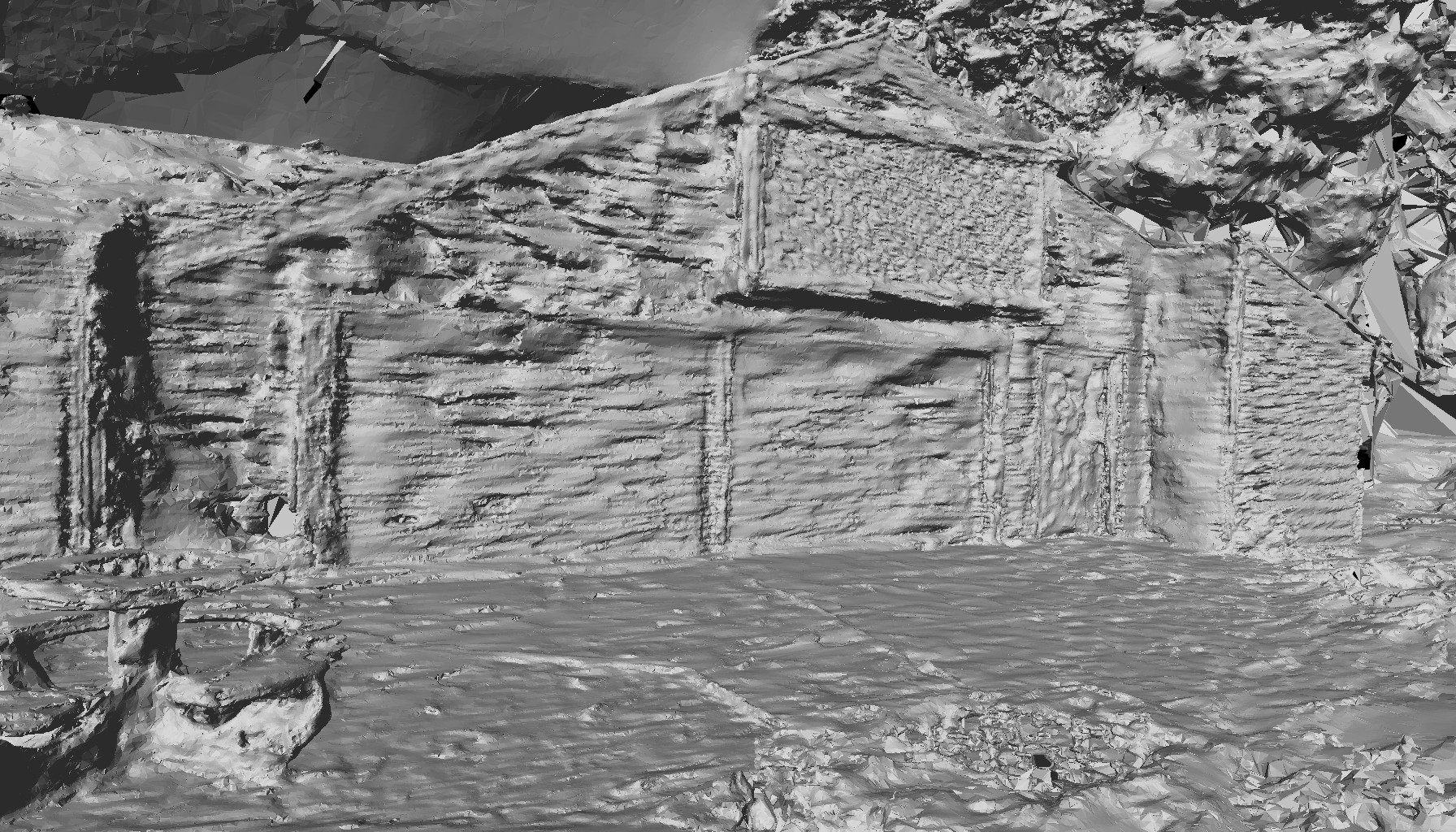}&
    \vcenterimg{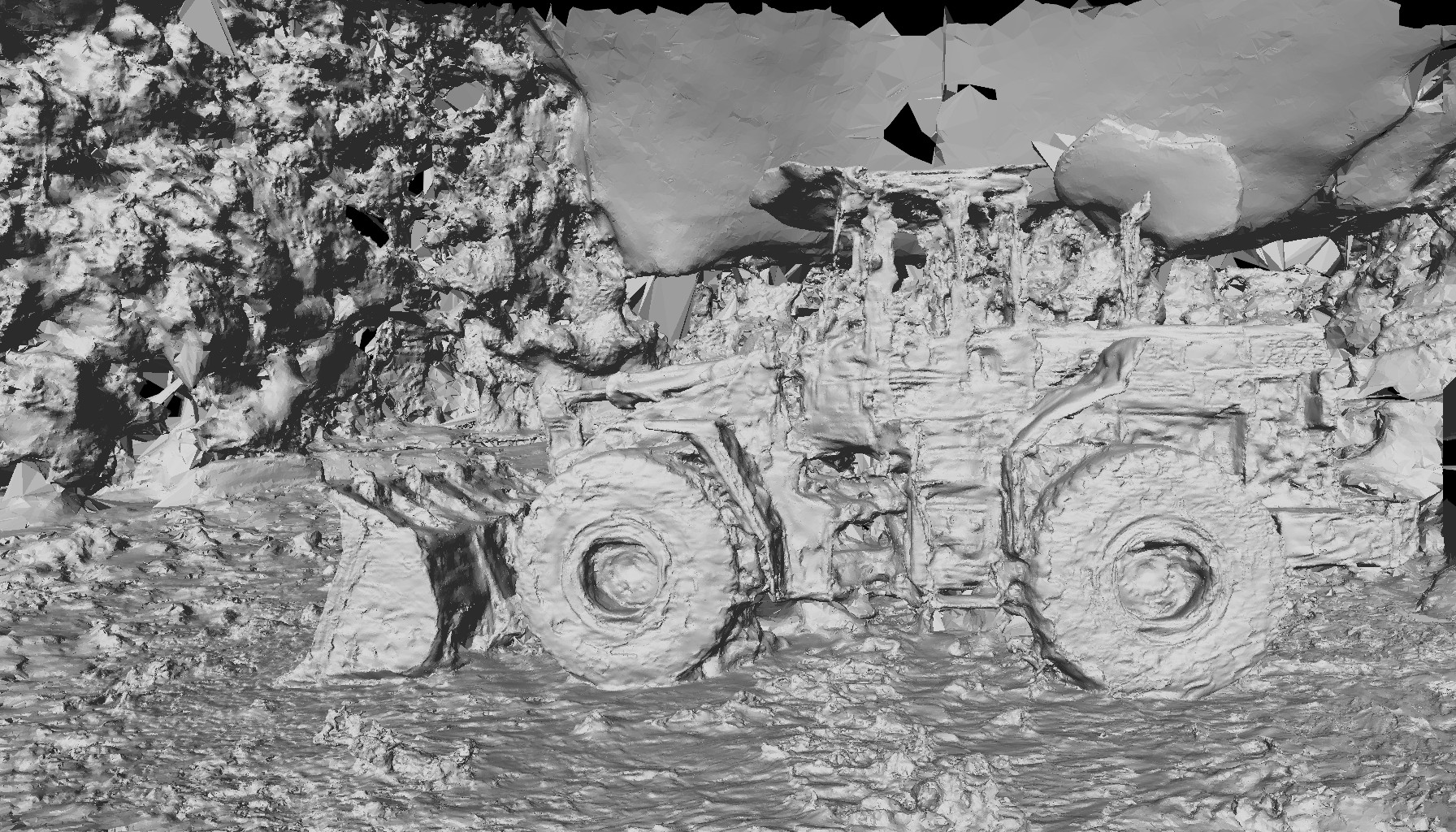}&
    \vcenterimg{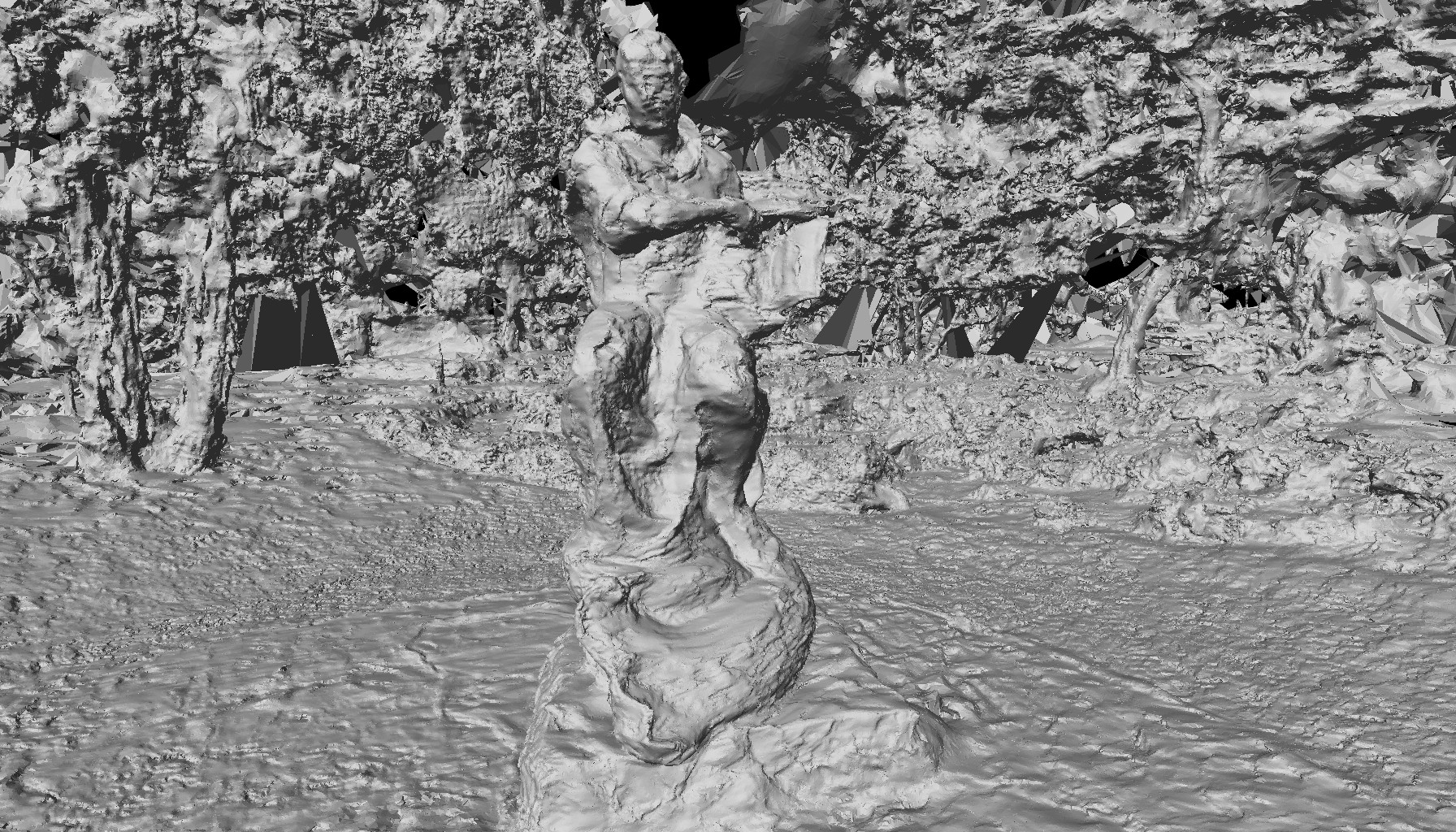}&
    \vcenterimg{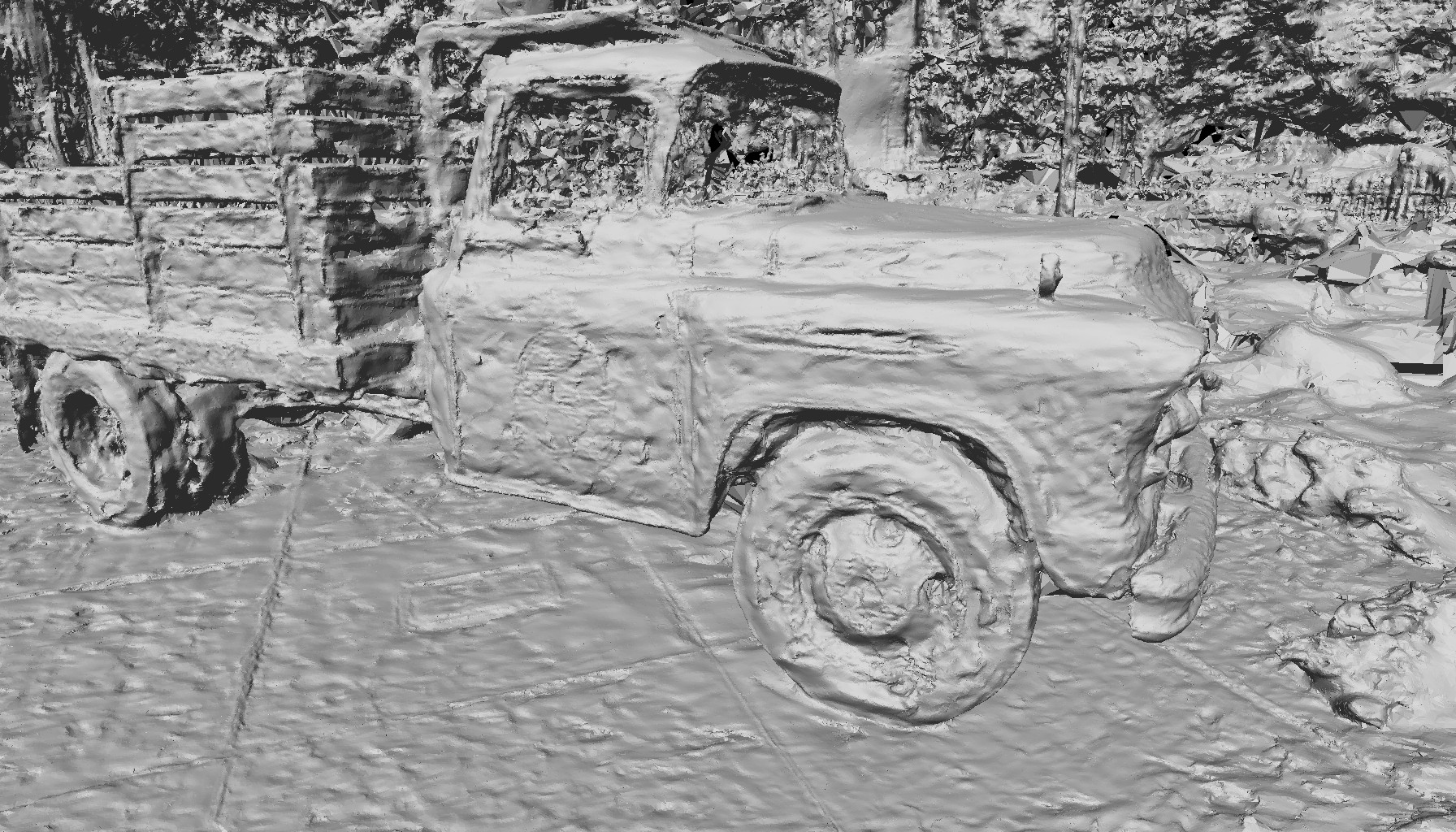}
    \end{tabular}
    
    \caption{\textbf{Mesh Reconstruction Results on the Tanks and Temples dataset.} Given 16 input views, our method produces the most accurate meshes against other feed-forward models. Moreover, while per-scene optimization methods overfit to sparse input images, our method yields noticeably smoother, cleaner, and less noisy surfaces.}
    
    \label{fig:MeshCompBench}
\end{figure*}

\begin{table}[ht]
\centering
\caption{\textbf{Photometric Quality.} We report visual quality metrics and the High-Resolution Rendering Consistency (HRRC)~\cite{he2026surfsplat} across different resolutions and sampling rates on novel views. Our method trained with the 3D smoothing filter, RS+GGGS+EANC and RS+GOF+EANC, excels at high resolutions since the predicted Gaussians do not exceed the Nyquist frequency set by the input images. GGGS Opt. 4k is shown for reference but excluded from \textbf{best}/\underline{second-best} highlighting.}
\label{tab:hrrc}
\fontsize{7}{11.5}\selectfont
\setlength{\tabcolsep}{4pt}
\resizebox{\linewidth}{!}{
\begin{tabular}{lccc|ccc|ccc|ccc}
\toprule
& \multicolumn{3}{c|}{256$\times$448 (\textbf{1x})} & \multicolumn{3}{c|}{512$\times$896 (\textbf{2x HRRC})} & \multicolumn{3}{c|}{1024$\times$1792 (\textbf{4x HRRC})} & \multicolumn{3}{c}{Average} \\
Method & PSNR$\uparrow$ & SSIM$\uparrow$ & LPIPS$\downarrow$ & PSNR$\uparrow$ & SSIM$\uparrow$ & LPIPS$\downarrow$ & PSNR$\uparrow$ & SSIM$\uparrow$ & LPIPS$\downarrow$ & PSNR$\uparrow$ & SSIM$\uparrow$ & LPIPS$\downarrow$ \\
\midrule
SurfSplat~\cite{he2026surfsplat} & 19.127 & 0.537 & 0.367 & 19.054 & 0.534 & 0.475 & 17.538 & 0.477 & 0.599 & 18.573 & 0.516 & 0.480 \\
RS+gsplat & \textbf{22.548} & \textbf{0.696} & \textbf{0.252} & \underline{23.095} & 0.708 & 0.345 & 22.911 & 0.718 & 0.416 & \underline{22.851} & 0.707 & 0.337 \\
RS+GGGS+EANC & \underline{22.437} & \underline{0.691} & \underline{0.257} & \textbf{23.120} & \underline{0.716} & \underline{0.311} & \underline{23.004} & \underline{0.741} & \underline{0.363} & \textbf{22.853} & \underline{0.716} & \underline{0.310} \\
RS+GOF+EANC & 22.283 & 0.687 & \underline{0.257} & 23.089 & \textbf{0.719} & \textbf{0.303} & \textbf{23.019} & \textbf{0.748} & \textbf{0.353} & 22.797 & \textbf{0.718} & \textbf{0.305} \\
\midrule
GGGS Opt. 4k & 23.163 & 0.767 & 0.198 & 23.437 & 0.758 & 0.259 & 23.409 & 0.768 & 0.313 & 23.336 & 0.764 & 0.256 \\
\bottomrule
\end{tabular}}
\end{table}

\subsubsection{Photometric Benchmark Comparisons.}
To compare visual quality, we adopt SurfSplat's~\cite{he2026surfsplat} High Resolution Rendering Consistency and evaluate renderings at the input resolution, and $\times 2$, $\times 4$ resolution on novel views. As seen in \cref{tab:hrrc}, at $256\times 448$, the encoders trained with GGGS and GOF trail behind. We attribute this to the constraints of the geometric loss. However, at higher resolutions, specifically at $\times 4$ scaling, the encoders perform better since the 3D smoothing filter limits the frequency of the predicted Gaussians to be below the Nyquist frequency. In \cref{fig:hrrc_4x_comparison}, for instance, RS+gsplat produces extremely fine Gaussians to capture the bench's bars and subtle details on the leaves around the stump and the tree's texture, while RS+GGGS+EANC and RS+GOF+EANC correctly limit the primitive's extent. Finally, SurfSplat fails to align its predicted Gaussians resulting in streaks most noticeable in the treehill scene.

\section{Conclusion}
\label{sec:conclusion}
We presented an encoder capable of successfully recovering accurate geometry in a single forward pass given sparse-view images of unbounded scenes. To achieve this, we introduce an edge-aware normal consistency loss and a fine-tuning strategy to leverage encoders previously trained for visual quality. Our edge-aware geometric loss successfully steers the encoder to predict Gaussians that lie flat on the scene's surfaces. Our fine-tuning strategy relies on a 3D smoothing filter, rooted in both signal processing and projective camera models, to simulate a view-space 2D filter by unprojecting and applying it to each predicted primitive. Experimental results on out-of-distribution outdoor scenes demonstrate that our proposed geometric loss and fine-tuned models are competitive with per-scene optimization methods while producing more compact meshes and outperform other feed-forward models in mesh reconstruction and high-resolution color rendering metrics.

\clearpage  %

\bibliographystyle{splncs04}
\bibliography{main}
\end{document}